\documentclass[letterpaper]{article} 
\usepackage{aaai25}  
\usepackage{times}  
\usepackage{helvet}  
\usepackage{courier}  
\usepackage[hyphens]{url}  
\usepackage{graphicx} 
\usepackage{natbib}  
\usepackage{caption} 
\usepackage{algorithm}
\usepackage{algorithmic}
\newcommand\blfootnote[1]{%
  \begingroup
  \renewcommand\thefootnote{}\footnote{#1}%
  \addtocounter{footnote}{-1}%
  \endgroup
}

\usepackage{newfloat}
\usepackage{listings}
\DeclareCaptionStyle{ruled}{labelfont=normalfont,labelsep=colon,strut=off} 
\floatstyle{ruled}
\newfloat{listing}{tb}{lst}{}
\floatname{listing}{Listing}
\title{Text2Relight: Creative Portrait Relighting with Text Guidance}
\author{
    Junuk Cha\textsuperscript{\rm 1,2$\dagger$}, Mengwei Ren\textsuperscript{\rm 2}, Krishna Kumar Singh\textsuperscript{\rm 2}, He Zhang\textsuperscript{\rm 2}, Yannick Hold-Geoffroy\textsuperscript{\rm 2}, Seunghyun Yoon\textsuperscript{\rm 2}, HyunJoon Jung\textsuperscript{\rm 2}, Jae Shin Yoon\textsuperscript{\rm 2*}, Seungryul Baek\textsuperscript{\rm 1*}
}
\affiliations{
    \textsuperscript{\rm 1} UNIST \quad \textsuperscript{\rm 2} Adobe Research
}

\usepackage{bibentry}

\begin{document}

\twocolumn[{
\renewcommand\twocolumn[1][]{#1}%
\maketitle
\begin{center}
    \captionsetup{type=figure}
    \includegraphics[width=\linewidth]{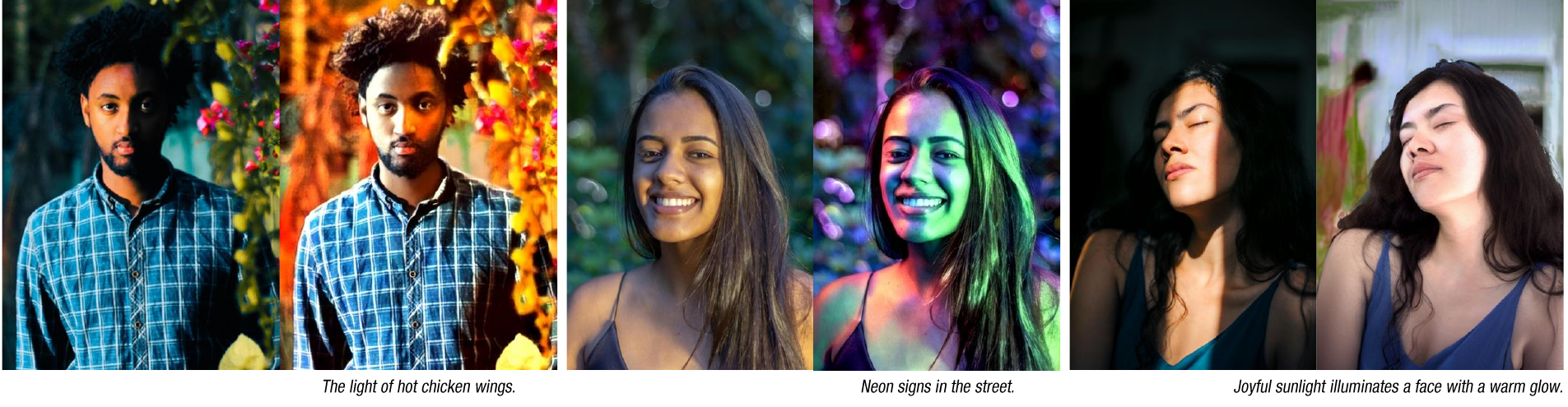}
    \caption{\textbf{Text2Relight} generates the image of a relighted portrait (right) as a condition of a text prompt while keeping original contents in an input image (left).}
    \label{fig:teaser}
\end{center}
}]

\blfootnote{\textsuperscript{$\dagger$}This research was conducted when Junuk Cha was an intern at Adobe Research. \textsuperscript{$*$}Co-last authors.\\
Copyright \copyright \ 2025, Association for the Advancement of Artificial Intelligence (www.aaai.org). All rights reserved.}

\begin{abstract}
We present a lighting-aware image editing pipeline that, given a portrait image and a text prompt, performs single image relighting. Our model modifies the lighting and color of both the foreground and background to align with the provided text description.
The unbounded nature in creativeness of a text allows us to describe the lighting of a scene with any sensory features including temperature, emotion, smell, time, and so on.
However, the modeling of such mapping between the unbounded text and lighting is extremely challenging due to the lack of dataset where there exists no scalable data that provides large pairs of text and relighting, and therefore, current text-driven image editing models does not generalize to lighting-specific use cases.
We overcome this problem by introducing a novel data synthesis pipeline:
First, diverse and creative text prompts that describe the scenes with various lighting are automatically generated under a crafted hierarchy using a large language model (\textit{e.g.,} ChatGPT). 
A text-guided image generation model creates a lighting image that best matches the text.
As a condition of the lighting images, we perform image-based relighting for both foreground and background using a single portrait image or a set of OLAT (One-Light-at-A-Time) images captured from lightstage system.
Particularly for the background relighting, we represent the lighting image as a set of point lights and transfer them to other background images.
A generative diffusion model learns the synthesized large-scale data with auxiliary task augmentation (\textit{e.g.,} portrait delighting and light positioning) to correlate the latent text and lighting distribution for text-guided portrait relighting.
In our experiment, we demonstrate that our model outperforms existing text-guided image generation models, showing high-quality portrait relighting results with a strong generalization to unconstrained scenes. 
Project page: https://junukcha.github.io/project/text2relight/
\end{abstract}

\section{Introduction}
Light, as a physical phenomenon, has been extensively studied in computer graphics to model its behavior as accurately as possible from real-world observations using a number of parameters such as color, intensity, and direction.
The advent of powerful generative AI technologies such as a denoising diffusion model~\cite{ho2020denoising} and large language models~\cite{radford2019language} have shifted the research paradigm from how \textit{accurately} to how \textit{creatively} one can model such a physical phenomenon.
Is it possible to model the lighting behavior as a function of our emotions? For example, how does a \textit{joyful} lighting look like? 
In this paper, we break the boundary between the physical and creative space by introducing Text2Relight, a lighting-specific foundational model that can perform relighting (\textit{i.e.,} remove the original lighting and apply a novel one) of a single portrait image driven by a text prompt as shown in Figure~\ref{fig:teaser}.
Our main assumption is that there exists unbounded creativeness in language, and correlating this with lighting enables the mapping from a physical space to a creative one.

\begin{figure}[t]
    \centering
    \includegraphics[width=1\linewidth]{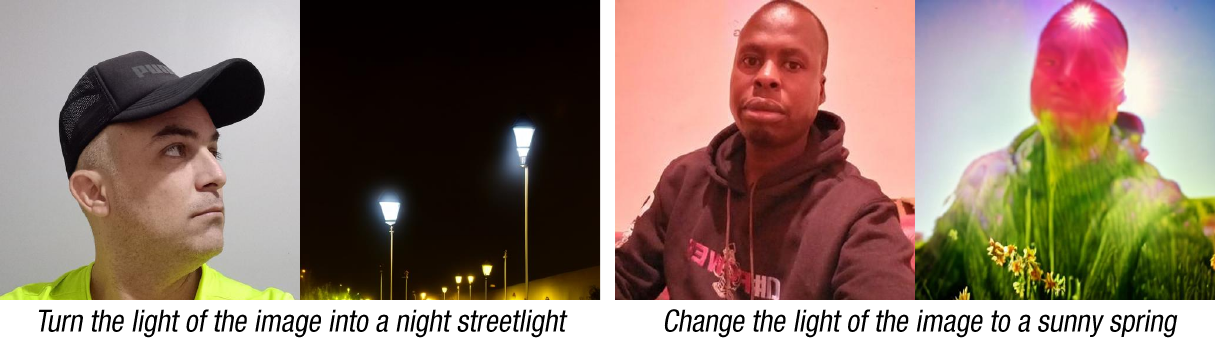}
    \caption{Results from existing text-guided image editing models (left: \cite{brooks2023instructpix2pix}, right: \cite{fu2023guiding}) which largely distorts the input images by generating new contents.}
    \label{Fig:motivation}
\end{figure}

One simple way to achieve Text2Relight is to utilize existing foundational models for text-guided image editing such as InstructPix2pix~\cite{brooks2023instructpix2pix}.
However, such a general foundational model which never learns from lighting-specific data, \textit{e.g.,} identical scenes captured under different lighting, encodes a weak correlation of the text with lighting, producing a large distortion of the original contents as shown in Figure~\ref{Fig:motivation}.
This necessitates a new foundational model that can control the image from the lighting space that is completely decomposed from the contents space (\textit{e.g.,} shape and intrinsic appearance).

However, developing such a lighting-specific model is challenging due to the lack of data pairs for relighting, \textit{i.e.,} the images of identical scene and main subject captured under different lighting conditions, associated by a text description.
While existing methods~\cite{pandey2021total,saito2023relightable} have captured the relighting data using expensive infrastructure such as lightstage system, such lab-controlled data are often not scalable (particularly for the axis of human identities) and the rendering of image relighting is often applicable to only foreground human region where the background scene is simply composed with a part of preset panorama images.
Those limited imaging data, in turn, restricts the diversity of the labeled text prompts as well.

To address this data challenge, we propose a scalable data simulation pipeline that can synthesize the relighting data of a portrait scene for both foreground and background, and associated text prompts.
Our synthesis pipeline is designed with bottom-up fashion: text generation, text-aware lighting image generation, and image-based relighting.
A large language model automatically generates diverse and creative text prompts based on our crafted language hierarchy to describe lighting environment of a scene.
Text-guided image generation models generate an RGB image or an HDR panorama map as a condition of a text prompt, which are used as a lighting image.
Finally, the lighting distribution of the generated lighting image is transferred to a portrait image using various image-based relighting methods; where a number of factors including data availability, algorithm maturity, and scene complexity lead us to approach the image-based relighting differently for background and foreground.

For foreground relighting, we develop an end-to-end portrait relighting model that can control the lighting of an input image as a function of a background image; but when OLAT (One-Light-at-A-Time) images are available from lightstage, we apply HDR rendering techniques using the generated panorama map. %
Since our method handles the case with only a single image, it can be applicable to any in-the-wild portrait scene.

For background relighting, we represent the scene lighting as a set of point lights~\cite{kocsis2023intrinsic} and introduce a robust and efficient lighting optimization method.
The positions of the point lights are initialized with a distance-based localization algorithm; and they are jointly optimized with other learnable variables (\textit{e.g.,} intensity and diffusion parameters) by minimizing the photometric difference from the lighting image. We relight a background image by transporting the optimized light sources using an inverse rendering techniques.

Using our large simulation data, we develop a lighting-specific foundational model by repurposing of an existing text-guided image editing model~\cite{brooks2023instructpix2pix}. 
In training time, the model jointly learns with an auxiliary task such as portrait shadow removal and text-guided light positioning to improve the geometric awareness and better intrinsic appearance modeling.

In our experiments, we show the analysis that our hierarchical text generation is indeed useful to push the distribution of the text diversity.
Our method outperforms existing text-guided foundational models in terms of pixel-wise perceptual distance, AI-based semantic score, and user preference score. 
We also provide in-depth ablation studies on the crafted language hierarchy, various data sources, and conditional variables to validate the effectiveness of our data simulation and model training pipeline.
As applications, we demonstrate portrait shadow removal, light positioning, and background harmonization.

In summary, our main contributions include:

\begin{itemize}
    \item Text2Relight, a new formulation of a lighting-specific foundational model for text-guided portrait relighting.
    \item A novel and scalable data simulation pipeline to synthesize relighted images and associated text prompts.
     \item A method to generate largely distributed light-aware text prompts with crafted hierarchy.
    \item Performance enhancement by joint training with auxiliary tasks (shadow removal and light positioning).
\end{itemize}

\section{Related Work}
\noindent\textbf{Portrait Relighting}
refers to the process of adjusting the lighting conditions in a portrait image including direction, intensity and color, so as to simulate different lighting scenarios, or creatively manipulate the lighting to achieve desired aesthetic effects.

Image-based human relighting has been extensively studied since it requires a minimum requirement (\textit{e.g.,} only a single image) for practical real-world application.
While many existing works have introduced a relighting model targeting various human body parts such as face, portrait, and full body, their fundamental approaches consist of two steps: intrinsic decomposition and shading from a target lighting.
Previous works first predict the intrinsic appearance of humans such as albedo, shading, and geometry from a single image using a neural network, and re-render the shading using graphics techniques such as HDR rendering from a panorama environment~\cite{pandey2021total,wang2023sunstage,yeh2022learning,ji2022geometry} and ray tracing~\cite{hou2022face}.
Some methods~\cite{hou2021towards,zhou2019deep} learn to generate shading as a function of a small number of spherical harmonics, and encoding the latent lighting rotation features onto the relighting model is also possible by learning from the augmented panorama environment maps with rotations~\cite{song2021half}. 
To improve physical plausibility, some methods predict more detailed intrinsic values such as roughness and reflectivity~\cite{kim2024switchlight} or predict more residual appearance~\cite{nestmeyer2020learning,tajima2021relighting,ji2022geometry} that can not be modeled by a simple intrinsic decomposition.
Conditioning explicit geometry such as 3D face model~\cite{zhou2019deep,ponglertnapakorn2023difareli} and 3D tri-plane~\cite{mei2024holo} in the latent space further helps to improve the geometric plausibility of the lighting behavior.
Data-driven lighting estimation from a user scribble~\cite{mei2023lightpainter} further enhances the controllability of the lighting.
Other than two-step relighting approaches, some works introduced a method that can directly change the lighting distribution from an image to improve the relighting efficiency.
A neural network jointly decodes the latent person image as well as the latent lighting parameters such as spherical harmonics~\cite{zhou2019deep}, latent background feature~\cite{ren2023relightful}, or latent illumination map~\cite{sun2019single}.


\noindent \textbf{Text-Guided Image Editing.}
Relevant to our task, recent advancements in diffusion models have significantly enhanced the capacity for sophisticated image editing directed by textual descriptions.  \cite{avrahami2022blended,nichol2021glide} enable precise local editing within specified regions of an image using masks, such as object replacement or object style transfer. The advantage of local editing is that it allows modification of the desired regions while preserving the rest of the image. In contrast, InstructPix2Pix~\cite{brooks2023instructpix2pix} globally edits the source image to match the text prompt. While it provides an overall natural-looking image, it sometimes modifies the context too aggressively, altering the identity of the portraits and/or the characteristics of the scenes. To address this issue, MGIE~\cite{fu2023guiding} utilizes a multimodal large language model to obtain more expressive instructions, which aids in precise local editing while making global adjustments. Imagic~\cite{kawar2023imagic} further enables complicated non-rigid editing by optimizing text embeddings aligned with the source and target images. 

Nevertheless, current text-based editing methods are primarily designed for general use cases, lacking specific considerations in modeling the lighting distribution that may fall beyond the existing large-scale training data. Consequently, they often fail when presented with relighting-specific prompts. Additionally, to our knowledge, there are no meticulously curated prompt-image pairs tailored for fine-tuning these models for the relighting tasks.

\section{Method}
\subsection{Problem Formulation}
Given a portrait image, our goal is to control the lighting of the scene for both foreground and background driven by a text prompt, while ensuring that the original content and identity are preserved: $\tilde{I} = f_{\theta}(I, M, T)$, where $\theta$ denotes the learnable parameters, and $f$ is the text-guided relighting function which takes as input source image $I$, foreground mask $M$, and text prompt $T$, and generates the relighted image $\tilde{I}$.

\begin{figure}
    \centering
    \includegraphics[width=1\linewidth]{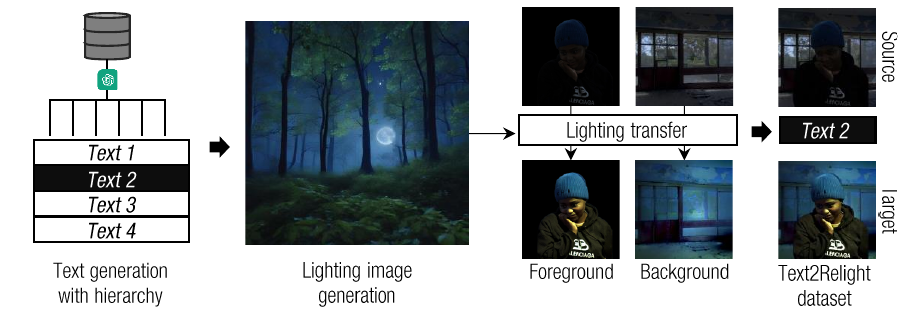}
    \caption{\textbf{Overview of the data synthesis pipeline.} We first generate a text prompt with a language hierarchy from which we generate a lighting image. Subsequently, we transfer the lighting from the lighting image to a portrait image captured from either lightstage or real world (with background inpainting). These form the training dataset for our Text2Relight model.}
    \label{fig:pipeline}
\end{figure}

To learn this mapping function, $f$ needs to be trained with the ground truth $\tilde{I}_{gt}$. However, no dataset provides pairs of the corresponding texts and relighted images that preserve the content and identity of the original images. To address those problems, we propose a novel dataset synthesis method consisting of three main components of text generation, text-driven lighting image generation, and image-based relighting as shown in Figure~\ref{fig:pipeline}.

We first generate diverse light-aware texts using our crafted language hierarchy. %
We use the generated text as a prompt to generate a lighting image in the form of either RGB image or HDR panorama map.
We transfer the lighting distribution from the generated lighting image to the foreground portrait and background scene, separately.
By compositing them together, we complete the synthesis of the relighted portrait image. 
Using the synthesized data, we develop a lighting-specific foundational model for Text-to-Relight with auxiliary task and data augmentation.
The details of each component is described in below.

\begin{figure}
    \centering 
    \includegraphics[width=1\linewidth]{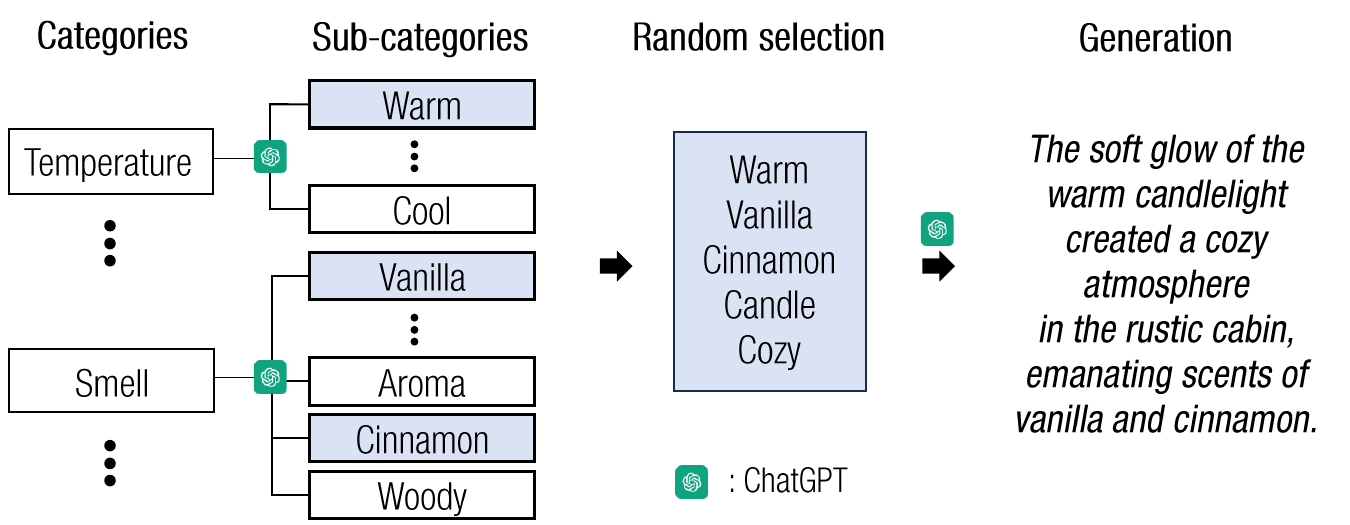}
    \caption{\textbf{Pipeline for text generation with hierarchy.} For sub-categories generation, we ask \textit{`Generate words related to} $\{$category$\}$. \textit{Write 30 or more words on a single line, separated by commas.'}; and for sentence generation, we ask \textit{`could you describe the lighting property of a random scene using the words of} $\{$selected words$\}$.' to ChatGPT, respectively.}
    \label{fig:pipeline_text}
\end{figure}

\subsection{Scalable Prompt Generation with Hierarchy}\label{sec:prompt}
We create large-scale text prompts that describe a scene in the context of lighting distribution using a large language model (LLM), \textit{e.g.,} ChatGPT~\cite{chatpgt2022openai}.
However, we empirically found that existing LLMs utilize a limited range of words to generate the text prompts from a simple question \textit{e.g., could you describe an arbitrary scene with its lighting distribution?}, which prevents us from generating diverse and creative text prompts.
Our approach, instead, is to explicitly select a few words from a predefined large vocabulary pool and provide them as a constraint on LLMs \textit{e.g., could you describe the lighting property of a random scene using the words of `cozy' and `warm'?}.
To define such a large vocabulary pool, we formulate a categorical hierarchy.
For example, humans define high-level categories related to lighting, and LLM generates various sub-categories for each category. 
The overall process for our hierarchical text generation is described in Figure~\ref{fig:pipeline_text}.
Please refer to the definition of category and sub-category definition in the supplementary. 

\subsection{Lighting Image Generation}
\label{sec:lighting_image_generation}
Given a text prompt, we generate two types of lighting images: RGB image and HDR panorama map as shown in Figure~\ref{fig:pipeline_foreground}. For RGB image, we employ a latent consistency model~\cite{luo2023latent} to generate the RGB image from a text prompt using four denoising steps. We use this image as the lighting image when we relight the foreground and background. For HDR panorama map, we develop a customized text-guided panorama generation model by fine-tuning a pre-trained stable diffusion model \cite{rombach2022high} on publicly available HDR panorama maps with paired text prompts extracted from an existing vision language model. It generates an HDR panorama map from a text prompt.

\subsection{Foreground Portrait Relighting}\label{sec:fore}
\begin{figure}
    \centering
    \includegraphics[width=1\linewidth]{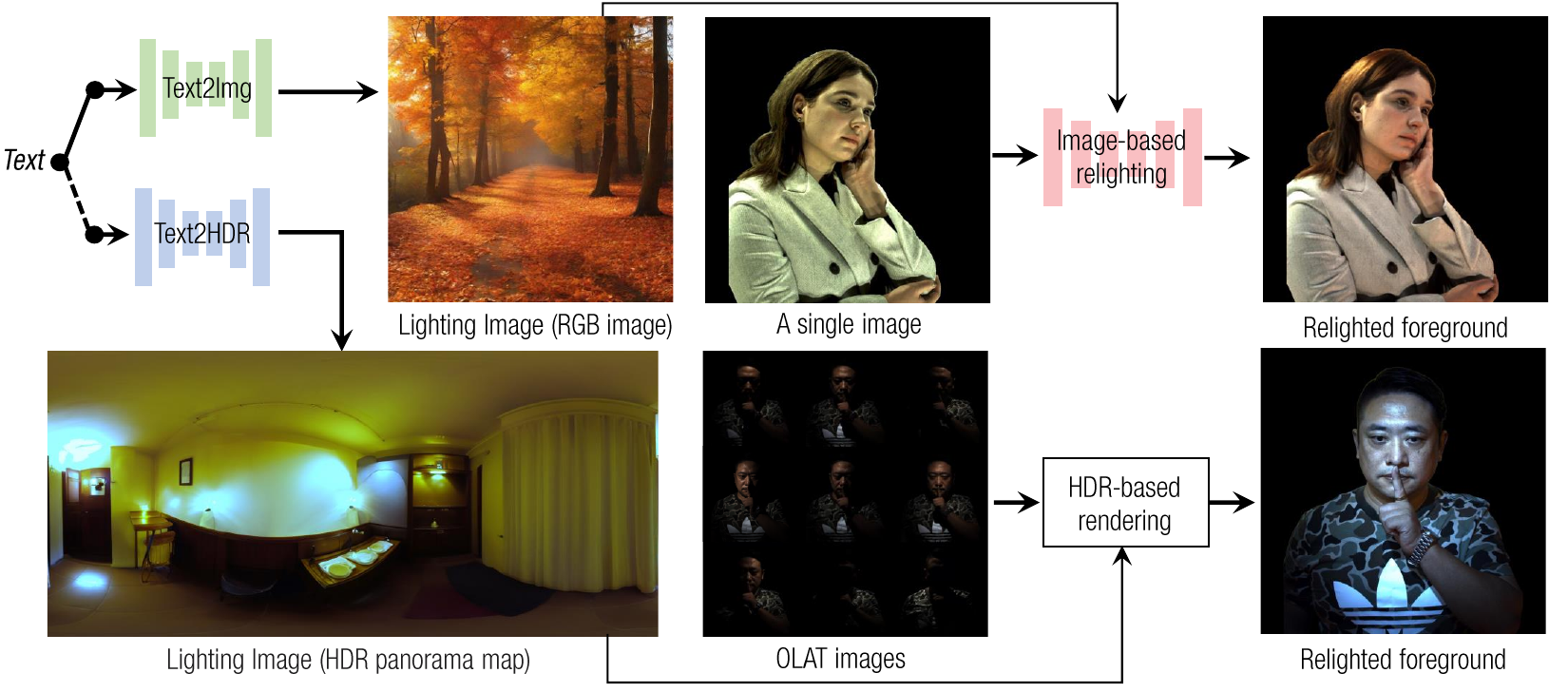}
    \caption{\textbf{Pipeline for foreground relighting}.}
    \label{fig:pipeline_foreground}
\end{figure}

We apply two separate methods for portrait relighting depending on the modality of the data: a single image and OLAT images. For a single-image-based portrait relighting, we develop an image-based relighting model that takes as input a single portrait image and a lighting image, and outputs a relighted image as shown in the upper row of Figure~\ref{fig:pipeline_foreground} using internal lightstage dataset and many real-world data similar to existing method~\cite{ren2023relightful}. For OLAT images captured from a lightstage as shown in the lower row of Figure~\ref{fig:pipeline_foreground}, we incorporate a panorama map as a lighting image to pre-defined the illumination condition for each OLAT image and perform the relighting of the portrait by applying HDR rendering techniques~\cite{zhang2021neural}.

\subsection{Background Relighting}
\label{sec:back}
Many existing works~\cite{careaga2023intrinsic,kocsis2023intrinsic} represent a background as multiple layers of intrinsic values: $I = A * S$ where $I$, $A$, and $S$ represent the background image, its albedo, and its shading, respectively. Furthermore, the shading can be described as a function of geometry and lighting: $S=s(L, G)$ where $s$ is a rendering function that outputs the shading map $S$ as a function of input lighting information $L$ and geometry $G$ which is often composed of depth $D$ and surface normal $N$, \textit{i.e.,} $G\rightarrow \{D, N\}$. Assuming $L$ is under the definition of a point lighting~\cite{iwahori1990reconstructing}, we can expand this to the multiple light sources as follows: $S=\sum^{n}_{i=1}S_i=\sum^{n}_{i=1}s(L_i, \{D, N\})$, where each $S_i$ corresponds to the shading contribution from an individual light $L_i$ and $n$ represents the number of point lights.

\begin{figure}
    \centering
    \includegraphics[width=1\linewidth]{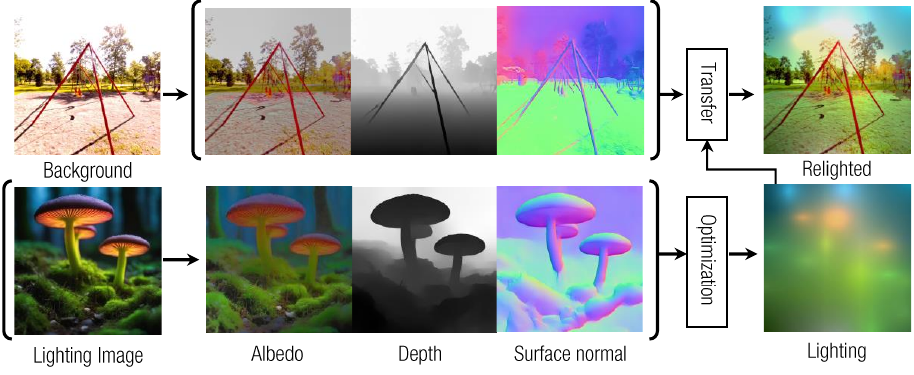}
    \caption{\textbf{Pipeline for background relighting}.}
    \label{fig:pipeline_background}
\end{figure}

Since lighting $L$ is completely decomposed from other intrinsic values, it is possible to transfer the lighting distribution to other background:

\begin{eqnarray}
    \tilde{I} = \hat{A}*\tilde{S}=\hat{A}*\sum^{n}_{i=1}s(L_i, \{\hat{D}, \hat{N}\}), 
    \label{eq:lighting_transfer}
\end{eqnarray}
where $\hat{\cdot}$ denotes the properties of the source background, and $\tilde{I}$ and $\tilde{S}$ are the relighted background and associated shading, respectively.

To reconstruct the point lights from an image, we first initialize a set of 3D point lights (\textit{e.g.,} 20) considering the intensity distribution of the gray-scaled image, and we optimize the parameters of each point light (\textit{i.e.,} 3D position, lighting intensity, ellipsoid ratio, and a diffusion intensity) based on the photometric errors between the reconstructed and the original lighting image. We transfer the optimized point lights to other background images using Eq.~\ref{eq:lighting_transfer}. Please refer to the supplementary material for more details about the point light reconstruction and transfer.

\subsection{Data and Task Augmentation}\label{sec:augment}
We further push the visual and lighting diversity of the simulated dataset with various data and task augmentation strategies. For data augmentation, we perform text prompt augmentation, spatial image augmentation, real-data augmentation, and bi-direction relighting augmentation. For task augmentation, we jointly our Text2Relight model using two different tasks: portrait shadow removal (\textit{i.e.,} remove portrait shadow or highlights while preserving original lighting environement) and light positioning (\textit{e.g.,} position a colored light around the portrait prompted by a text). Please refer to the supplementary materials for more details about our data augmentation and the way to create the dataset for task augmentation. 

\begin{figure}[t]
    \centering
    \includegraphics[width=0.98\linewidth]{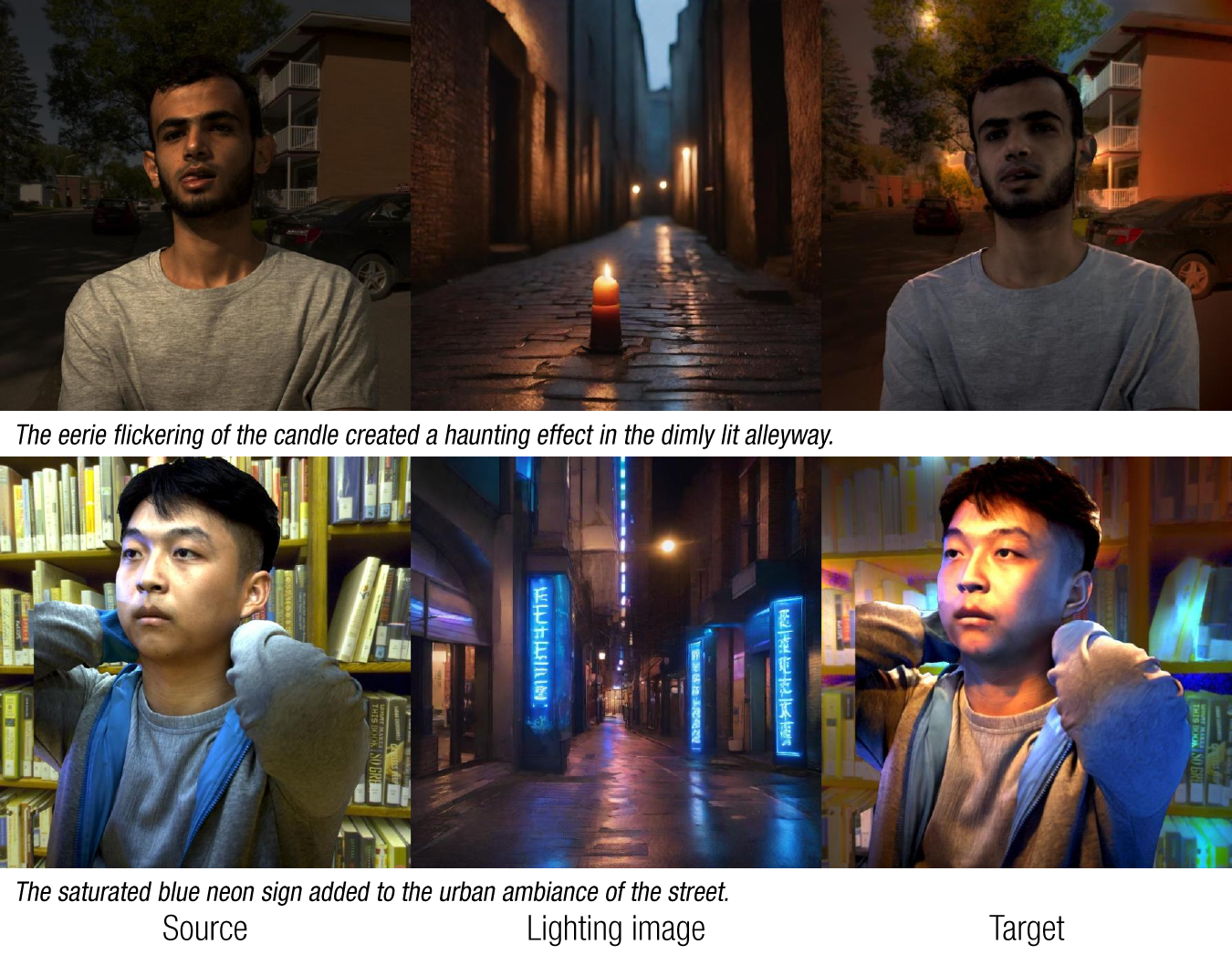}
    \caption{Examples of our simulated data. Source, lighting image, target and associated text prompt. Source and target have same content but different lighting environment.}
    \label{fig:examples_data}
\end{figure}

\subsection{Model Training}\label{sec:train}
We repurpose a text-guided image editing model (\textit{i.e.,} InstructPix2Pix~\cite{brooks2023instructpix2pix}) as a lighting-specific foundational (Text2Relight) model by learning the following objectives:
\begin{eqnarray}
    \mathcal{L}_{T2R}(x) = || \epsilon - f_\theta(\{z_{t}, I, M\}, t, T)(x) ||^2_2.
\end{eqnarray}
$x$ is the latent pixel position, $z_{t}$ is the intermediate noisy latent at time $t$, $f_\theta$ is the denoiser (\textit{i.e.,} UNet) which predicts the noise, $T$ is text, $I$ and $M$ are portrait image and foreground mask, respectively. $M$ guides the model to focus on the foreground, helping it learn effectively and prevent to generate lighting artifacts. 

\noindent\textbf{Dataset Summary.} Overall, our data has 1.5M pairs of relighting images and associated text prompts. We create 400K and 800K data from OLAT images and a single image, respectively. We create 100K data pair for shadow removal from a lightstage and 200K data pair for light positioning using a single image. The examples of our simulated data are shown in Figure~\ref{fig:examples_data}.

\begin{table}[t]
    \centering
    \caption{Comparison with other state-of-the-art methods. CVS, FIS, and LS denote CLIP vision score, face identity score, and LLaVA score, respectively.}
    \begin{tabular}{lccccc}
    \hline
    Method & SSIM $\uparrow$ & LPIPS $\downarrow$ & CVS $\uparrow$ & FIS $\uparrow$ & LS $\uparrow$ \\
    \hline
    IP2P & 0.408 & 0.531 & 0.646 & 0.008 & 3.6 \\
    GLIDE & 0.464 & 0.525 & 0.643 & -0.969 & 3.6 \\
    MGIE & 0.415 & 0.464 & 0.802 & 0.465 & 3.3 \\
    Ours & \textbf{0.546} & \textbf{0.401} & \textbf{0.886} & \textbf{0.584} & \textbf{3.8} \\
    \hline
    \end{tabular}
    \label{tab:sota}
\end{table}

\section{Experiments}

\noindent\textbf{Datasets.}
For quantitative evaluation, we use our data simulation pipeline to synthesize the ground-truth data for text-guided portrait relighting. We use real-world portrait images from \cite{kvanchiani2023easyportrait} as testing sets.
For qualitative evaluation, we use many real-world portrait images collected from license-free stock data.

\noindent\textbf{Metrics.}
We utilize various metrics to measure the score between the generated images and ground truths. SSIM~\cite{wang2004image} measures structural similarity between two images. LPIPS~\cite{zhang2018perceptual} evaluates perceptual similarity between two images by comparing their deep features from a neural network~\cite{simonyan2014very}. The CLIP vision score (CVS) calculates the cosine similarity between two images using CLIP vision encoder~\cite{radford2021learning}. To validate how effectively the identity of the portrait is preserved, we use the face identity score (FIS) that computes the cosine similarity between two face images in the feature embedding space constructed using ArcFace~\cite{deng2019arcface}. To measure the score of how well the text prompt matches an image, we fine-tune the LLaVA~\cite{liu2024visual} using the data obtained from ChatGPT~\cite{chatpgt2022openai}: a question template, a concatenated image, and an answer of ChatGPT. To this end, we create the concatenated image by merging a source image and an output image in parallel and input them into the ChatGPT to get the score of the lighting adjustment on a scale from 1 (very bad) to 5 (very good). The fine-tuned LLaVA measures the lighting adjustment score from 1 to 5. The details of the question template used in our validation are shown in the supplemental material.

\noindent\textbf{Baselines.}
We compare our model with IP2P~\cite{brooks2023instructpix2pix}, GLIDE~\cite{nichol2021glide}, and MGIE~\cite{fu2023guiding} in Table~\ref{tab:sota}. For the fair comparison, we use an instruction-based prompt template `Change the portrait under lighting of \textit{\{Text prompt\}}' for existing models since they tend to distort the content when using \textit{\{Text prompt\}} as the text prompt.

\subsection{Comparison}
IP2P~\cite{brooks2023instructpix2pix} globally edits the source image based on the text prompt, and it sometimes dramatically changes the foreground of the portrait. GLIDE~\cite{nichol2021glide} employs a mask to mark an edited region. It preserves the content in the region outside the mask, but the content inside the mask region is dramatically changed. MGIE~\cite{fu2023guiding} leverages a large multi-modal model to obtain an expressive text prompt. However, it does not ensure the preservation of the foreground contents. According to face identity score (FIS) values, IP2P and GLIDE tend to change the face of the portrait. In particular, GLIDE largely changes the human to other objects. MGIE has the larger FIS than IP2P and GLIDE, but it sometimes creates artifacts on the foreground. Although the LLaVA score (LS) does not show significant differences, the qualitative results indicate that only our model can create new shadows and lighting. Existing models merely perform style transfer rather than actual relighting. The detailed qualitative comparisons are shown in Figure~\ref{fig:qualitative}. More results are illustrated in the supplemental material. 

\begin{figure}[t]
    \centering
    \includegraphics[width=0.98\linewidth]{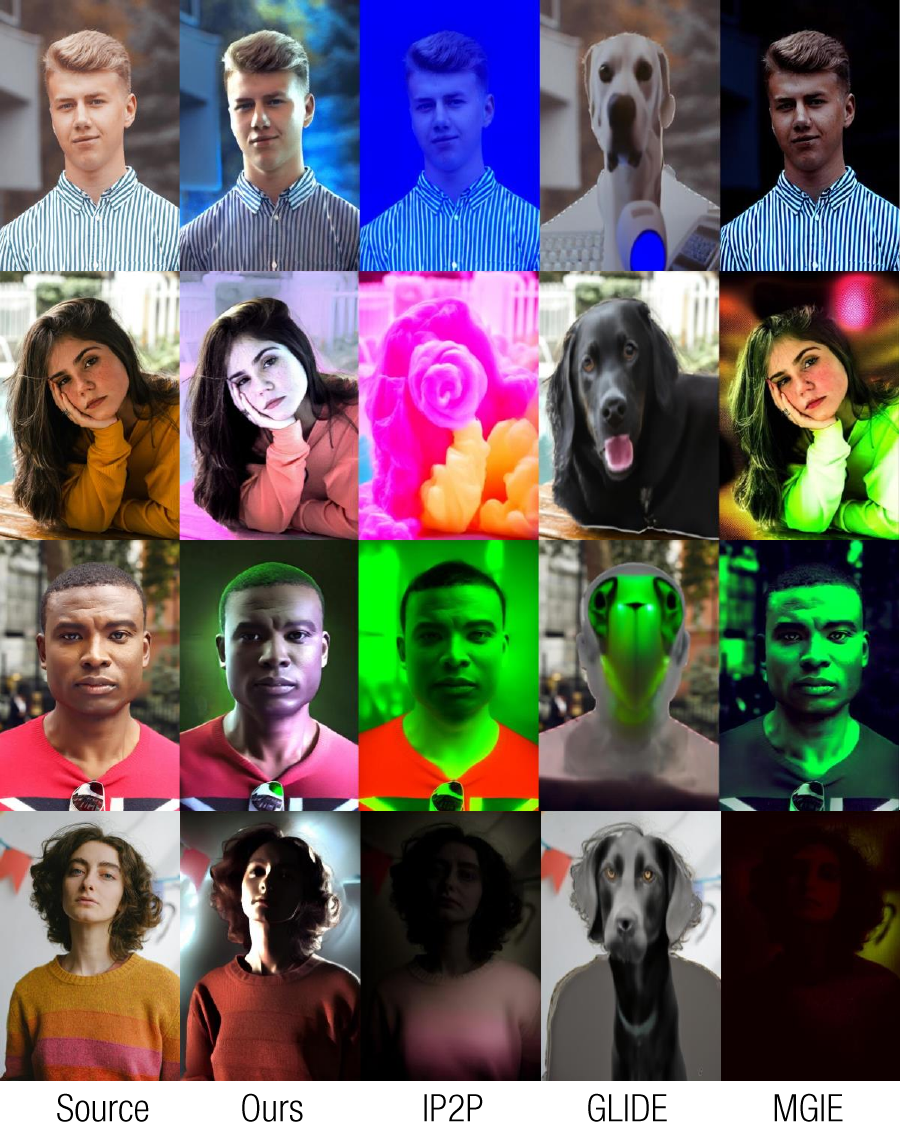}
    \caption{Qualitative comparison with text prompts : `The blue light of the computer monitor', `Sweet cotton candy', `Eerie glow of green fluorescent lighting', and `The chilling atmosphere of the fear-inducing dark room'. We compare ours with IP2P~\cite{brooks2023instructpix2pix}, GLIDE~\cite{nichol2021glide}, and MGIE~\cite{fu2023guiding}.}
    \label{fig:qualitative}
\end{figure}

\begin{table}[t]
    \centering
    \caption{User study for preference. 2.33\% select `None'.}
    \begin{tabular}{l|ccccc}
    \hline
    Method  & Ours & IP2P & MGIE & GLIDE \\
    \hline
    Preference & 66.17\% & 14.83\% & 15.50\% & 1.17\% \\
    \hline
    \end{tabular}
    \label{tab:user_study}
\end{table}

\begin{figure}[t]
  \begin{center}
    \includegraphics[width=0.85\linewidth]{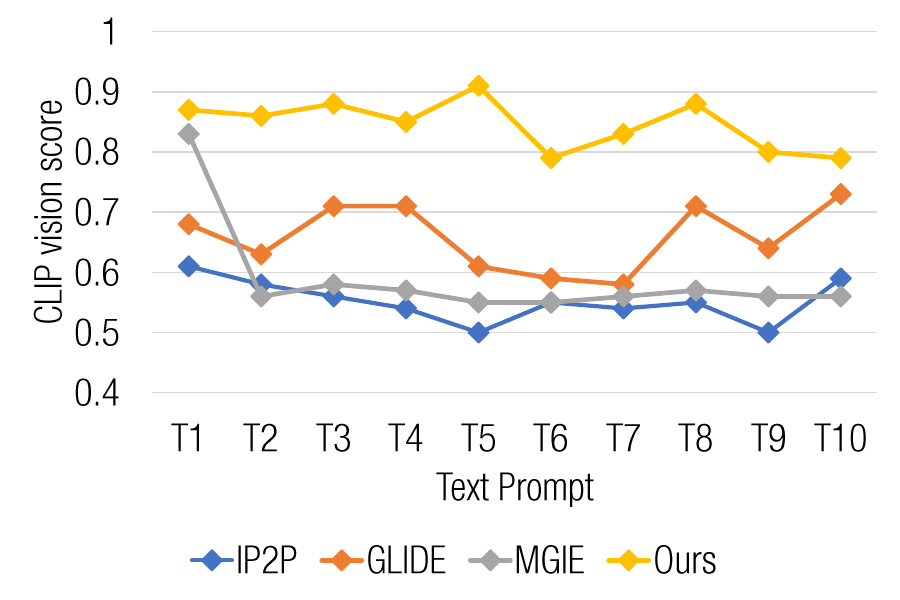}
  \end{center}
  \caption{Performance comparison according to 10 text prompts (T1 $\sim$ T10).}
  \label{fig:consitency_text_prompts}
\end{figure}

We further compare the CLIP vision score performance according to 10 text prompts, as shown in Figure~\ref{fig:consitency_text_prompts}. Our model consistently performs well across all text prompts. MGIE's performance varies depending on the text, while other models show low performance for all text prompts.

\begin{table}[t]
    \centering
    \caption{Ablation study based on data and conditions. ATD means the auxiliary task data.}
    \begin{tabular}{llccc}
    \hline
    Type & Method & SSIM $\uparrow$ & LPIPS $\downarrow$ & CVS $\uparrow$ \\
    \hline
    Data & Ours w/o ATD & 0.545 & 0.419 & 0.866 \\
    Condition & Ours w/o mask & 0.535 & 0.423 & 0.865 \\
    \hline
    Full & Ours & \textbf{0.551} & \textbf{0.399} & \textbf{0.882} \\
    \hline
    \end{tabular}
    \label{tab:ablation}
\end{table}

\subsection{User Study}
We conduct the user study to compare the user perceptual preference, involving 30 participants and 20 samples. We present users with randomly arranged images from various models: Ours, IP2P~\cite{brooks2023instructpix2pix}, MGIE~\cite{fu2023guiding}, and GLIDE~\cite{nichol2021glide}, and ask the following question: which portrait relighting result best matches the `text' while preserving the `content'? Table~\ref{tab:user_study} shows that our model's results are the most preferred, although some users occasionally choose images that contain the new contents described in the text. For details of the user study, please refer to the supplemental material.

\subsection{Ablation Study}
We conduct an ablation study on data, condition, and crafted hierarchy.

\begin{figure}[t]
    \centering
    \includegraphics[width=1\linewidth]{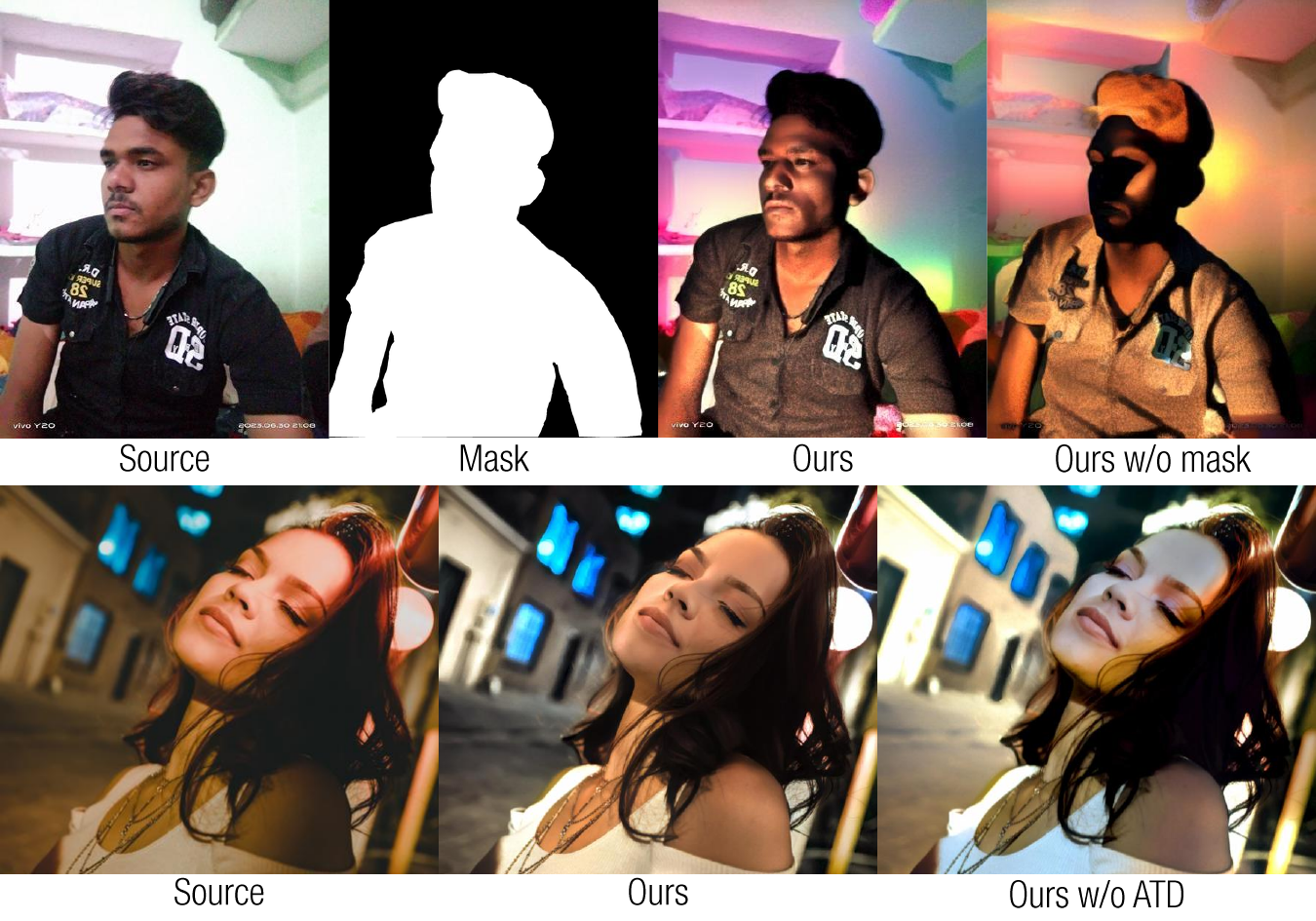}
    \caption{Ablation study on mask condition and auxiliary task data (ATD). We use text prompts, `The eerie and mysterious glow of the light added an enchanting touch to the dark forest' and `Sunlight illuminates the her face', respectively.}
    \label{fig:ablation}
\end{figure}

\noindent\textbf{Effect of Auxiliary Task Augmentation.} We utilize auxiliary task data (\textit{e.g.,} shadow-free and geometry lighting data). Table~\ref{tab:ablation}-(\textit{Data}) summarizes the study of the effect of the auxiliary task augmentation. \textit{Ours w/o ATD} means the model is trained without the auxiliary task data (ATD). Auxiliary task data helps with modeling the intrinsic appearance of humans under various shadows and highlights (with the delighting task), and the light positioning enables better geometric understanding, both of which lead to performance improvement. It is also presented in Figure~\ref{fig:ablation}.

\noindent\textbf{Mask Condition.} We use a mask condition for the Text2Relight model to distinguish the foreground and background. Table~\ref{tab:ablation}-(\textit{Condition}) shows the comparison of the trained model with and without mask conditioning. When the mask is not conditioned, the model sometimes generates the lighting artifacts, such as saturation and blur in the foreground, leading to poor performance. This issue is also confirmed in the qualitative results, as illustrated in Figure~\ref{fig:ablation}.

\begin{figure}[t]
    \centering
    \includegraphics[width=0.7\linewidth]{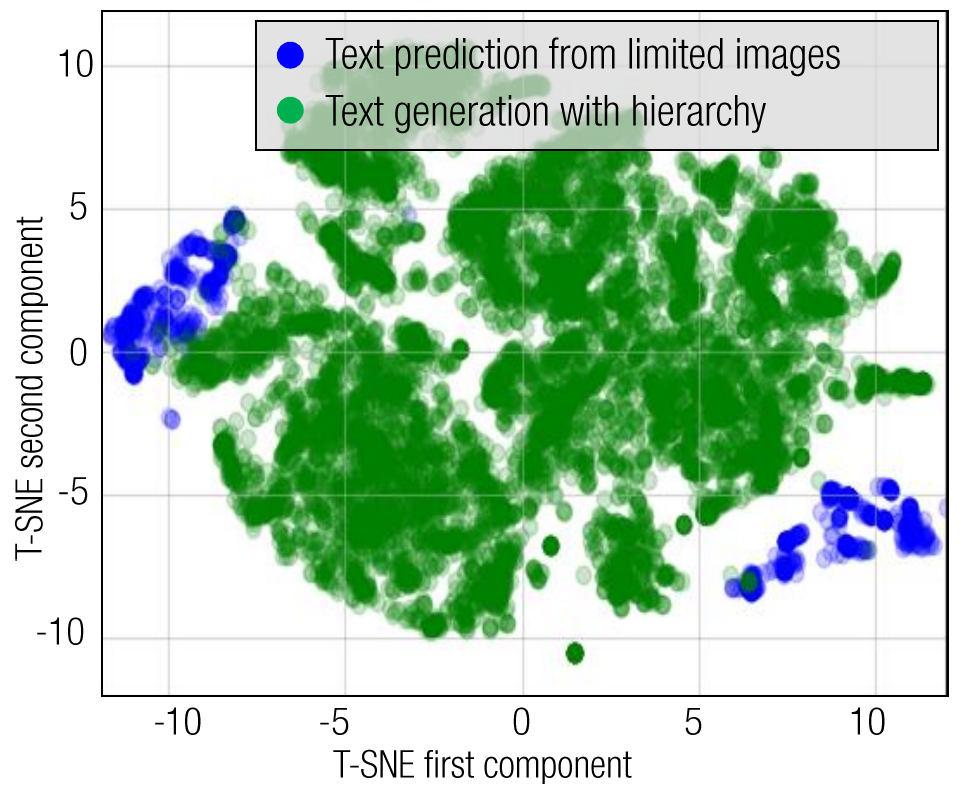}
    \caption{The T-SNE distribution of the texts predicted from limited sets of environment images using a vision language model (blue); and generated from a large-language model using our crafted language hierarchy (green).}
    \label{fig:text_distribution}
\end{figure}

\noindent\textbf{Crafted Language Hierarchy.} As shown in Figure~\ref{fig:text_distribution}, our crafted language hierarchy significantly enhances the large language model's ability to produce a wider variety of text prompts, surpassing the distributional limitations typically observed in text generated by vision-language models. This strategic enhancement ensures that our simulated data achieves greater diversity.

\subsection{Application}
\begin{figure}[t]
    \centering
    \includegraphics[width=0.98\linewidth]{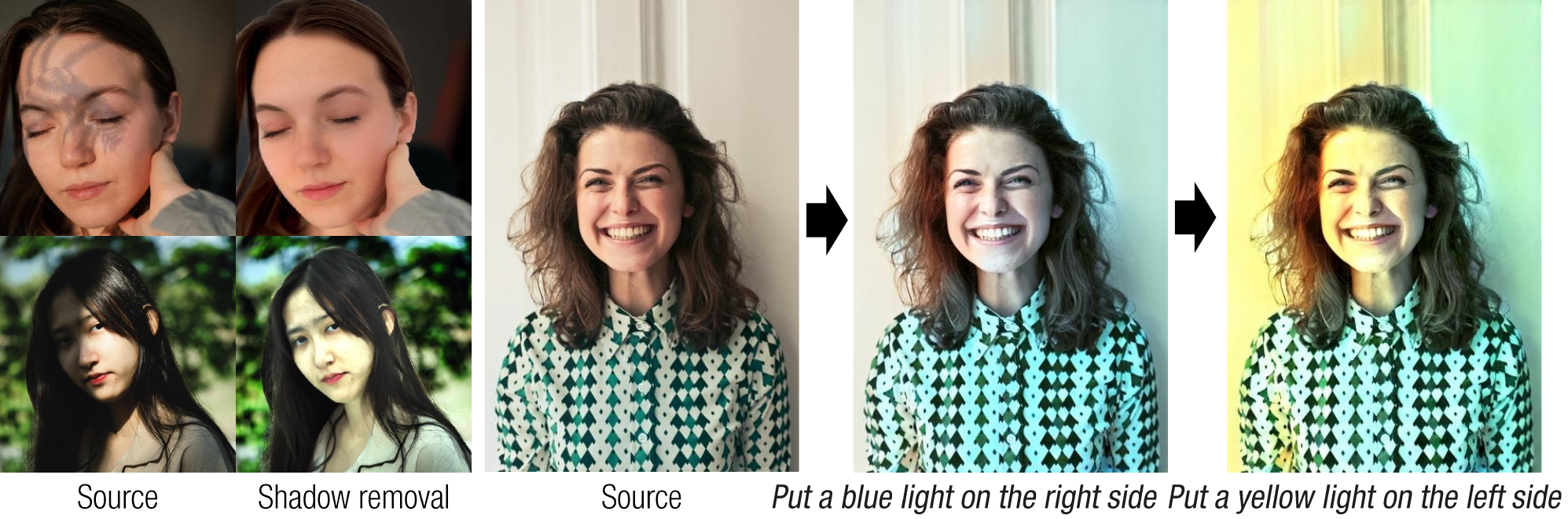}
    \caption{Examples of shadow removal and lighting position. First, shadow removal examples are demonstrated in the left two rows. Second, light positioning is shown in the right three images.}
    \label{fig:application_shadow_positioning}
\end{figure}

\noindent\textbf{Shadow Removal.} Our model can support portrait shadow removal. For example, in the left two rows of Figure~\ref{fig:application_shadow_positioning}, we use text prompts, `Eliminate the shadow from the portrait' and `Remove the shadow', for the shadow removal. 

\noindent\textbf{Light Positioning.} Our model can support the light positioning with a coarse text description about the light position. In Figure~\ref{fig:application_shadow_positioning}, we perform a series of text-guided relighting: `Put a blue light on the right side', and `Put a yellow light on the left side'. 

\begin{figure}[t]
    \centering
    \includegraphics[width=0.98\linewidth]{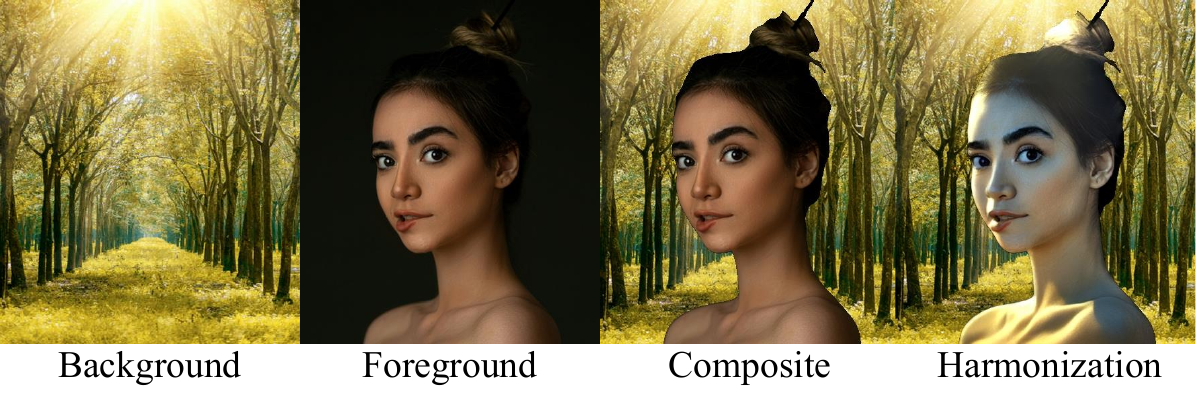}
    \caption{Examples of background harmonization.}
    \label{fig:application_bg_harmonize}
\end{figure}

\noindent\textbf{Background Harmonization.} We can use Text2Relight for background harmonization. For example, we apply the text prompt, `Natural relighting', to the initially composited portrait image with a novel background, and re-composite the relighted foreground and the intact background. As shown in Figure~\ref{fig:application_bg_harmonize}, the model relights the foreground considering the lighting distribution of the background.

Please refer to the supplementary for additional applications (\textit{e.g.,} general object relighting and latent interpolation).

\section{Conclusion}
We introduce Text2Relight, an end-to-end relighting model that can control the lighting distribution of a single image. 
We address the core dataset challenge by introducing a novel data simulation pipeline that can synthesize scalable ground-truth pairs of the relighting images and associated text prompt in three steps: generating diverse light-aware text prompts with our crafted language hierarchy, text-conditioned lighting image generation, and image-based foreground and background relighting.
We perform repurposing of a pre-trained diffusion model as a lighting-specific foundational model by learning our large-scale simulation data.
In our experiments, we demonstrate that Text2Relight effectively changes the lighting distribution that well-reflects the semantics of text prompts while maintaining the original contents from the input image, outperforming existing text-guided image editing models.

\noindent \textbf{Limitation.}
Our model sometimes generates some strong point lights in the background, which appear unnatural. With weak spatial lighting context (accurate 3D position and color information) from the nature of text, our model is sometimes confused to localize the text-specified lighting.

\section{Acknowledgments.}
Junuk Cha and Seungryul Baek were supported by IITP grants (No. RS-2020-II201336 Artificial intelligence graduate school program (UNIST) 25\%; No. RS-2021-II212068 AI innovation hub 25\%; No. RS-2022-II220264 Comprehensive video understanding and generation with knowledge-based deep logic neural network 50\%), funded by the Korean government (MSIT).

\bibliography{aaai25}

\twocolumn[{
\begin{center}
\textbf{\Large Text2Relight: Creative Portrait Relighting with Text Guidance\\-Supplementary-\\}
\vspace{16mm}
\end{center}
}]

\setcounter{equation}{0}
\setcounter{figure}{0}
\setcounter{table}{0}
\setcounter{page}{1}
\makeatletter
\renewcommand{\theequation}{S\arabic{equation}}
\renewcommand{\thefigure}{S\arabic{figure}}
\renewcommand{\thetable}{S\arabic{table}}


\section{Implementation Notes}
We implement our model in Pytorch~\cite{paszke2019pytorch} and train it using 8$\times$40GB A100 GPUs with a batch size of 1,024 for 512x512 resolution images. We initialize the UNet with pre-trained weights of IP2P~\cite{brooks2023instructpix2pix}.

\section{Details of Category and Sub-category}
\noindent\textbf{Category Definition and Sub-category Generation.} We define the following 19 categories that could be explicitly and implicitly related to the lighting behavior: atmosphere, color, temperature, directionality, emotions, intensity, light location, lighting effect, place, purpose of lighting, shape, smell, sound, source type, taste, time, touch, universe, and weather. Subsequently, we generate more than 30 sub-categories for each category using ChatGPT~\cite{chatpgt2022openai}. Please see Section `Text Examples' of the supplementary for the full sub-category definition.

\noindent\textbf{Sub-category Random Selection.} We randomly select two to six sub-categories as constraints on text generation. 
we assign higher weights to specific categories directly related to the physical behavior of the lighting such as position and color to ensure the model sufficiently learns physical correctness during training.

\section{Details of Point Light Reconstruction and Transfer}
\begin{figure}[t]
    \centering
    \includegraphics[width=1\linewidth]{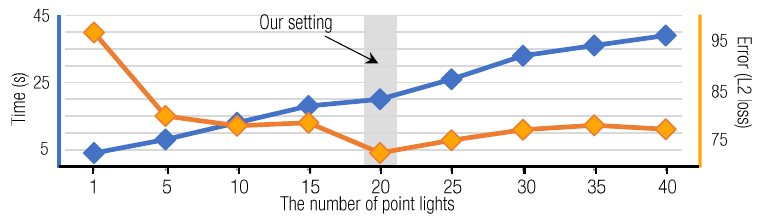}
    \caption{Graph of time and error that depend on the number of point lights for lighting optimization. Adding more lights increases the computational time and decreases the errors. However, adding too many light sources rather leads to the overfitting to a local minima, increasing the errors.}
    \label{fig:num_lights}
\end{figure}

\noindent\textbf{The number of Point Light.} In practice, we allocate 20 point lights (the ideal numbers in our experiments as described in Figure~\ref{fig:num_lights}).

\noindent\textbf{Point Lights Reconstruction by Optimization.}
Given a lighting image $I$, we reconstruct point lights by optimizing the following objective:
\begin{eqnarray}
    \mathcal{L} = || I - A*\sum^{n}_{i=1}s(L_i, \{D, N\})||^2_2,
\end{eqnarray}
where we obtain an albedo $A$ from $I$ using existing detection method \cite{careaga2023intrinsic}, and a normal $N$ and a depth $D$ are also detected from $I$ using internal detection models (please refer to `Monocular Surface Normal and Depth Detection module.' for more details). Each point light $L$ is composed of a set of learnable parameters including color $C$, 3D position $P=(x_L, y_L, z_L)$, intensity $\mathcal{I}$, ellipsoid ratio $\varepsilon$, and a diffuse parameter $\sigma$. By taking those parameters, the differentiable rendering function $s$ renders the shading at each pixel position $\{x,y\}$ under Lambertian reflectance~\cite{koppal2020lambertian} as follows:
\begin{eqnarray}
    s(L, \{D, N\},\{x,y\}) = \nonumber \\
    C \cdot [\mathcal{I} \cdot N(x, y) \cdot l(x, y, D(x, y); P, \varepsilon, \sigma)],
    \label{eq:point_light}
\end{eqnarray}
where
\begin{eqnarray}
    l(x, y, z; P, \varepsilon, \sigma) = \nonumber \\
    \frac{\left( x_L - x, \varepsilon \cdot (y_L - y), z_L - z \right)}{\left( \sqrt{((x_L - x))^2 + (y_L - y)^2 + (z_L - z)^2} \right)^\sigma},\nonumber
\end{eqnarray}
and $N(x, y)$ and $D(x, y)=z$ represent the surface normal and the depth at pixel position $\{x,y\}$, respectively.

In our experiments, how we initialize the 3D position of each point light largely affects the robustness and computational time. We allocate 20 point lights within a normalized 3D cube.
Their initial positions are configured using a distance-based selection algorithm which maximizes the minimum distance between the selected points that belong to the strong pixel intensity. This ensures that the 3D point lights are localized around the pixels with strong intensity while they are spreading each other. The depth of each point light is initialized using $D(x,y)$. For the initialization of other variables, we set the color as $(0.5, 0.5, 0.5)$, the intensity as $\frac{1}{\#L}$ where $\#L$ denotes the number of the lights, ellipsoid ratio as 1, and diffuse parameter as 1.

\noindent\textbf{Lighting Transfer.} We transfer the reconstructed point lights to the source background as described in Eq.~1 of the main paper: $\tilde{I} = \hat{A}*\sum^{n}_{i=1}s(L_i, \{\hat{D}, \hat{N}\})$ where $\hat{\cdot}$ denotes the properties of the source background, and $\tilde{I}$ presents the relighted background. However, directly transferring the lighting to the background often makes the lighting invisible, due to the depth differences between the background and lighting image. To address this issue, we transfer the relative distance between the scene (\textit{i.e.,} depth) and lighting position while keeping the same values for other parameters.

\noindent\textbf{Monocular Surface Normal and Depth Detection module.} $N$ and $D$ are also detected from $I$ using internal detection models which are composed of Unet architecture with pyramid vision transformer~\cite{wang2022pvt} that learns many mixtures of ground-truth data similar to~\cite{ranftl2020towards}. 

\begin{figure}[t]
    \centering
    \includegraphics[width=1\linewidth]{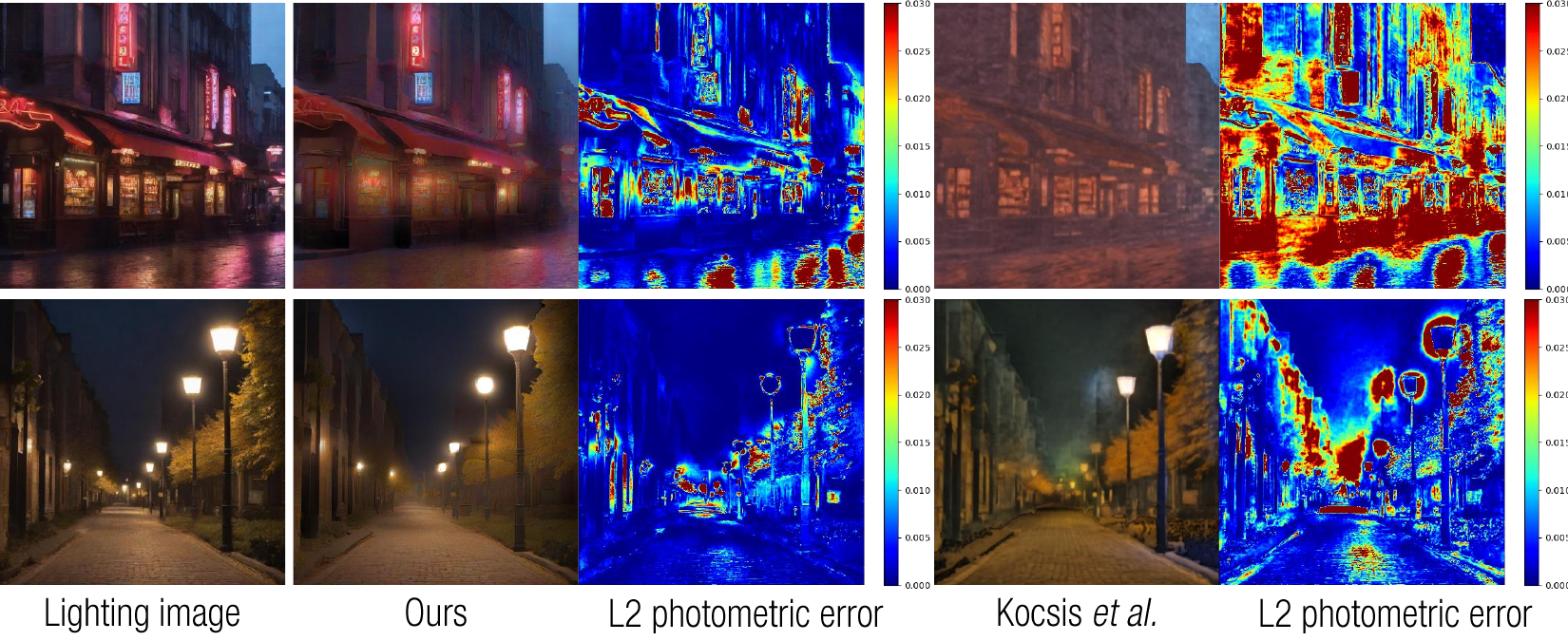}
    \caption{Qualitative example for the comparison of our lighting optimization method with ~\cite{kocsis2023intrinsic}. On average, the L2 photometric error of ours is $6.4 \times 10^{-3}$ with 38s processing time, and for Kocsis's method $16.5 \times 10^{-3}$ with 102s.}
    \label{fig:comparison_light_optim}
\end{figure}

\section{Data and Task Augmentation}
We further push the visual and lighting diversity of the simulated dataset with various data and task augmentation strategies. For effective generalization, beyond the conventional spatial image augmentation such as rotation, cropping, and padding, we perform real-data augmentation using existing portrait images~\cite{kvanchiani2023easyportrait}, bi-directional data training, and background contents augmentation of the source image by generation. For better modeling of intrinsic human appearance and geometric plausibility of the lighting behavior, we perform task augmentation by training our Text2Relight model jointly with two auxiliary tasks: shadow removal and light positioning.

\noindent\textbf{Text Prompt Augmentation.} We use various LLMs including Llama2~\cite{touvron2023llama} and LLaVA~\cite{liu2024visual}. Llama2 can modify words and structures within the original prompts to maintain their length as well as rephrase them to produce shorter prompts. Similar to ~\cite{fu2023guiding}, given a source image and a text prompt, LLaVA generates an expressive prompt with multi-modal capability that provides more specific details about the source image.

\noindent\textbf{Real World Images.} To enhance the model's generalizability, we use 20k real world images~\cite{kvanchiani2023easyportrait} as source images. To generate corresponding target images, we use the image-based data generation method as described in Sections `Foreground Portrait Relighting' and `Background Relighting' of the main paper. To separate the foreground and background from the source image, we use a foreground mask detector, which is composed of Unet architecture with pyramid vision transformer~\cite{wang2022pvt} that learns many masks of ground-truth data similar to~\cite{ranftl2020towards}. We then in-paint the foreground region using the fine-tuned latent diffusion model~\cite{rombach2022high} to generate a complete background.

\noindent\textbf{Swapping Source and Target for Data Augmentation.}
For Bi-directional data generation, we augment the data by swapping the source and target images. To obtain appropriate text prompts for swapped data, it is essential to accurately describe the lighting in the original source image. For this, we utilize LLaVA~\cite{liu2024visual}, which is fine-tuned using data from GPT-4-vision~\cite{achiam2023gpt}, to generate the required text prompt.

\noindent\textbf{Source Image Augmentation.}
For background contents augmentation of the source image, we use the HDR panorama map generated by Text2HDR~\cite{rombach2022high} and the image generated by Text2RGB~\cite{luo2023latent} as the source background. For HDR-based rendering, the data processing follows the method described in Sections `Foreground Portrait Relighting' and `Background Relighting' of the main paper; however, we use the generated HDR panorama map instead of the captured HDR panorama map to create the source image. For image-based relighting, the original portrait's foreground is relighted using the image-based relighting model conditioning on the generated background. Subsequently, we composite the relighted foreground and the generated background to produce the source image.

\noindent\textbf{Shadow Removal.} We collect the shadow removal data from a lightstage following existing portrait delighting works~\cite{futschik2023controllable} where we use diffused HDR panorama maps to render shadow-free portrait image from OLAT data, and associated texts are largely augmented in the context of shadow removal using Llama2~\cite{touvron2023llama}.

\noindent\textbf{Light Positioning.} Following a point lighting theorem~\cite{koppal2020lambertian} as also described in Eq.~\ref{eq:point_light}, we synthesize the single portrait image with extra point lights. We divide the image into nine grid sections and assign associated categories \textit{e.g.,} top-right, center, and so on. At each grid, we randomly sample the 3D position of a point light from a random distance. We pick a color from 100 preset categories. We put/move/remove maximum three point lights to the original image using Eq.~\ref{eq:point_light}, where we use the detected surface normal and depth. The associated text prompts are largely augmented with Llama2~\cite{touvron2023llama} given a template sentence sampled from many predefined action categories. Please refer to Section `Text Examples' for the full definition of our action and color categories.

\section{Comparison of Point Lights Reconstruction}
We compare our lighting optimization performance with \cite{kocsis2023intrinsic}, as shown in Figure~\ref{fig:comparison_light_optim}. Our method reconstructs and spreads various colored point lights better than Kocsis~\textit{et al.}'s method~\cite{kocsis2023intrinsic}. We measure the L2 photometric error, which quantifies the difference in pixel values between the lighting image and the reconstructed image. Our method achieves a lower error compared to Kocis~\textit{et al.}'s method. Additionally, our method (38s) is about three times faster than Kocsis~\textit{et al.}'s method (102s), which is indeed important for scale data generation. For the detailed animation process of our optimization, please refer to the supplemental video.

\begin{figure}[t]
    \centering
    \includegraphics[width=1\linewidth]{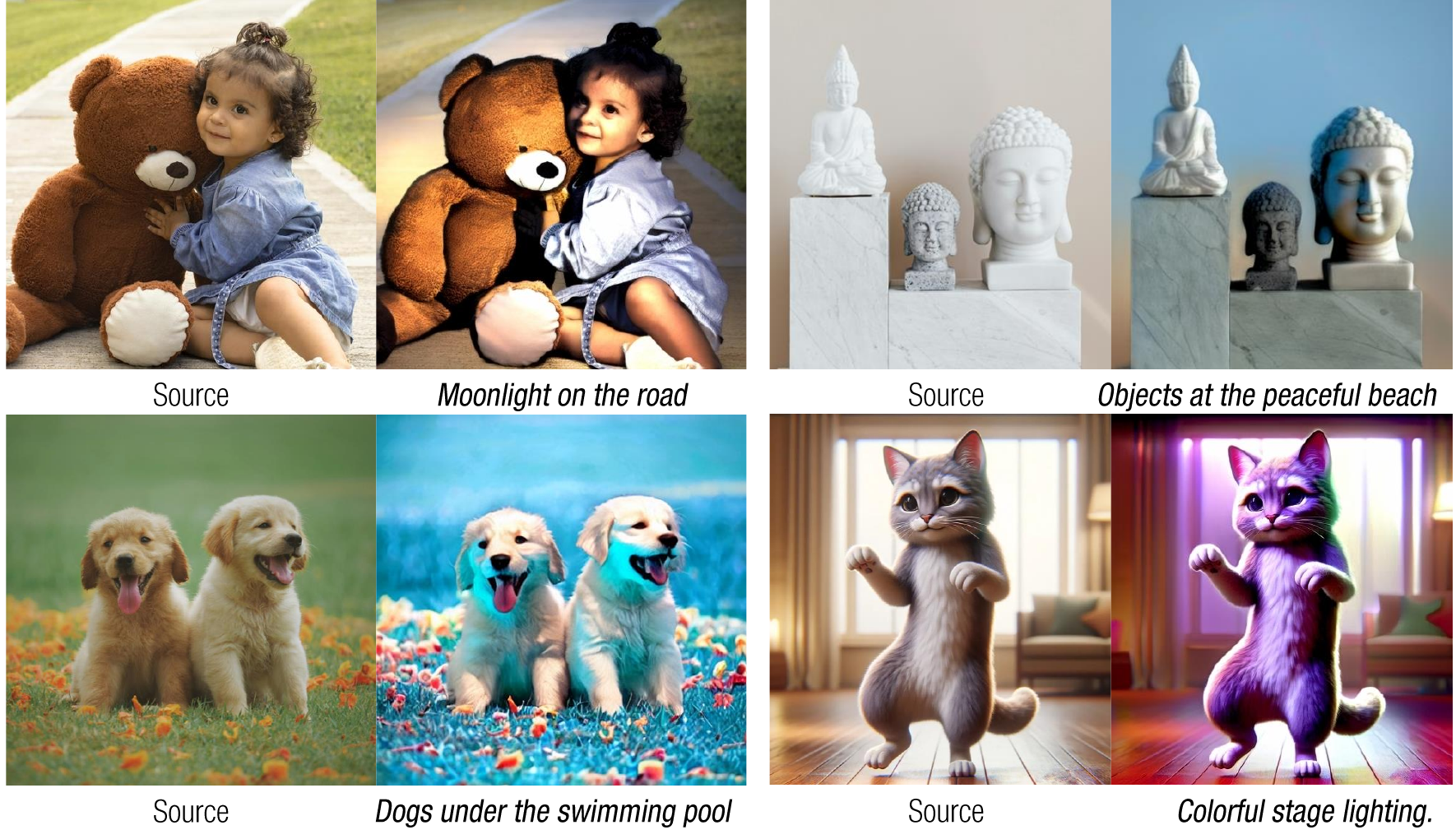}
    \caption{Examples of non-portrait images.}
    \label{fig:application_non_portrait}
\end{figure}

\begin{figure}[t]
    \centering
    \includegraphics[width=1\linewidth]{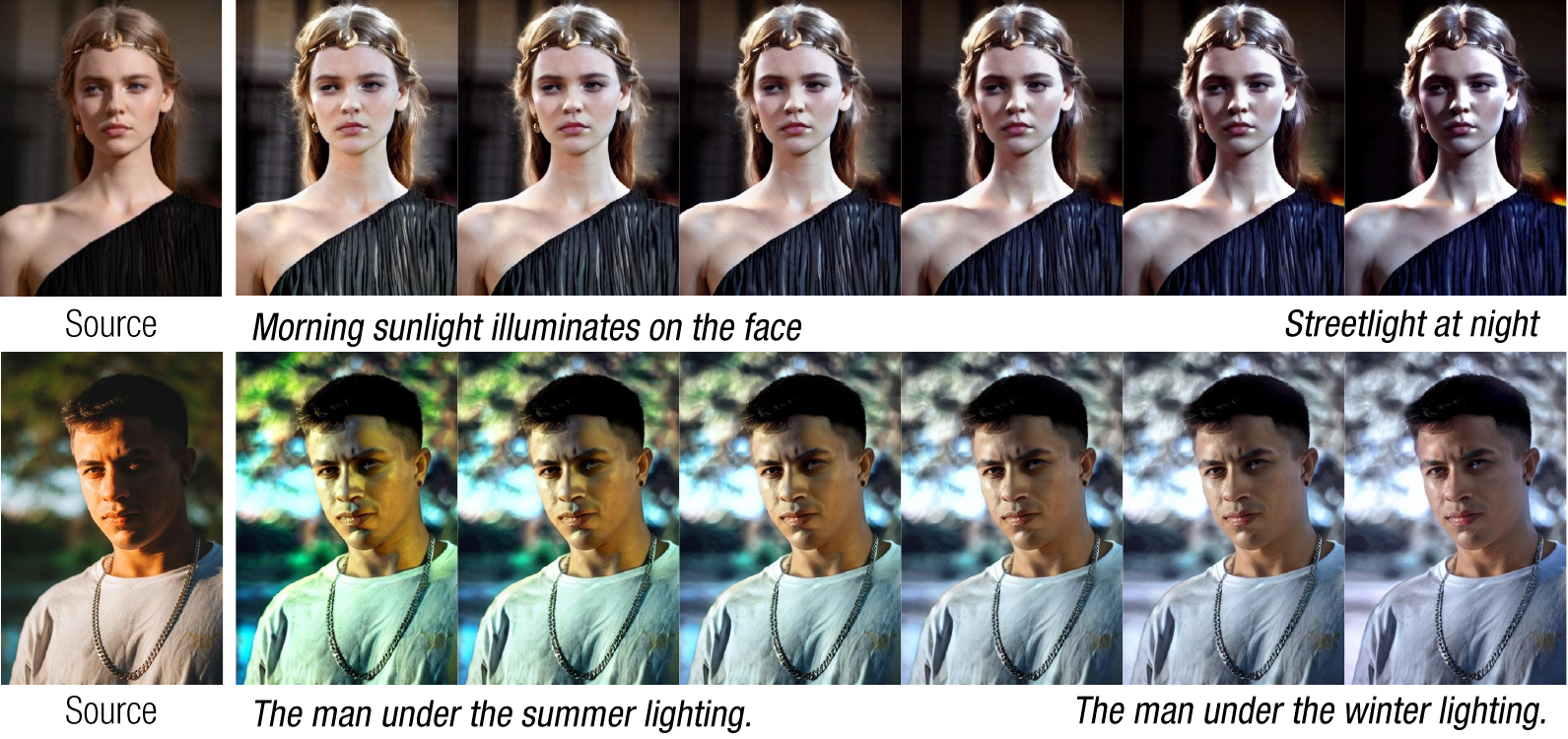}
    \caption{Examples of latent interpolation.}
    \label{fig:application_latent}
\end{figure}

\section{Application}
\begin{figure}[t]
    \centering
    \includegraphics[width=1\linewidth]{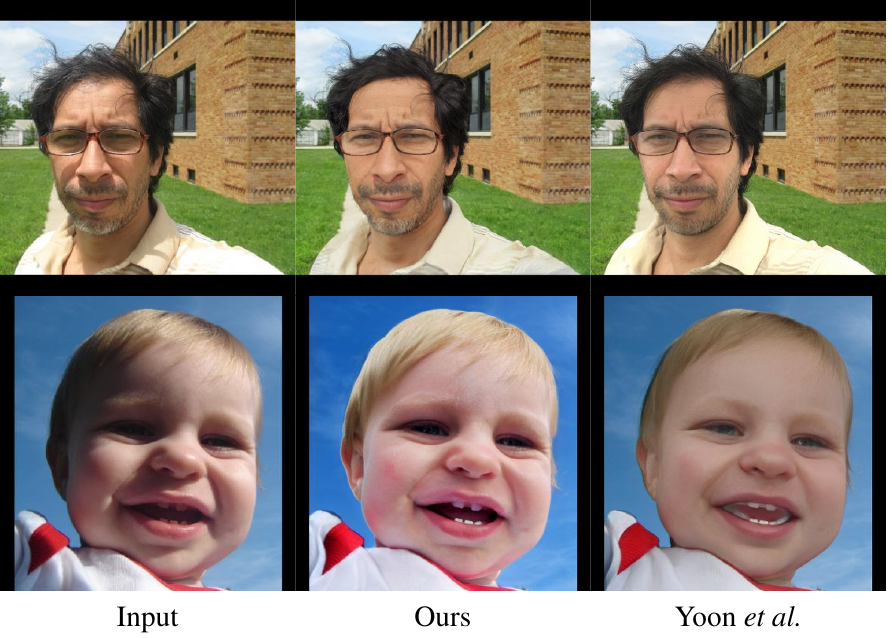}
    \caption{Comparison of shadow removal results between our model and the model proposed by Yoon~\textit{et al.}~\cite{yoon2024generative}}
    \label{fig:application_shadow_removal}
\end{figure}

\noindent\textbf{Shadow Removal.} We compare our model's performance in removing shadows from portraits. As shown in Figure~\ref{fig:application_shadow_removal}, our model demonstrates performance comparable to that of the model proposed by Yoon~\textit{et al.}~\cite{yoon2024generative}, which is specifically designed for shadow removal.

\noindent\textbf{Non-portrait Images.} Interestingly, our model is generalizable to non-portrait images thanks to the utilization of the pretrained large image priors. Figure~\ref{fig:application_non_portrait} shows some examples where the relighting results are highly reflective of the physical implication such as a novel shadow made by self-occlusion.

\noindent\textbf{Creative Latent Interpolation.} Similar to previous work \cite{song2020denoising}, we could demonstrate the late interpolation between the results from our creative relighting as shown in Figure~\ref{fig:application_latent}. More results and animations are shown in the supplemental video.

\begin{figure}[t]
    \centering
    \includegraphics[width=1\linewidth]{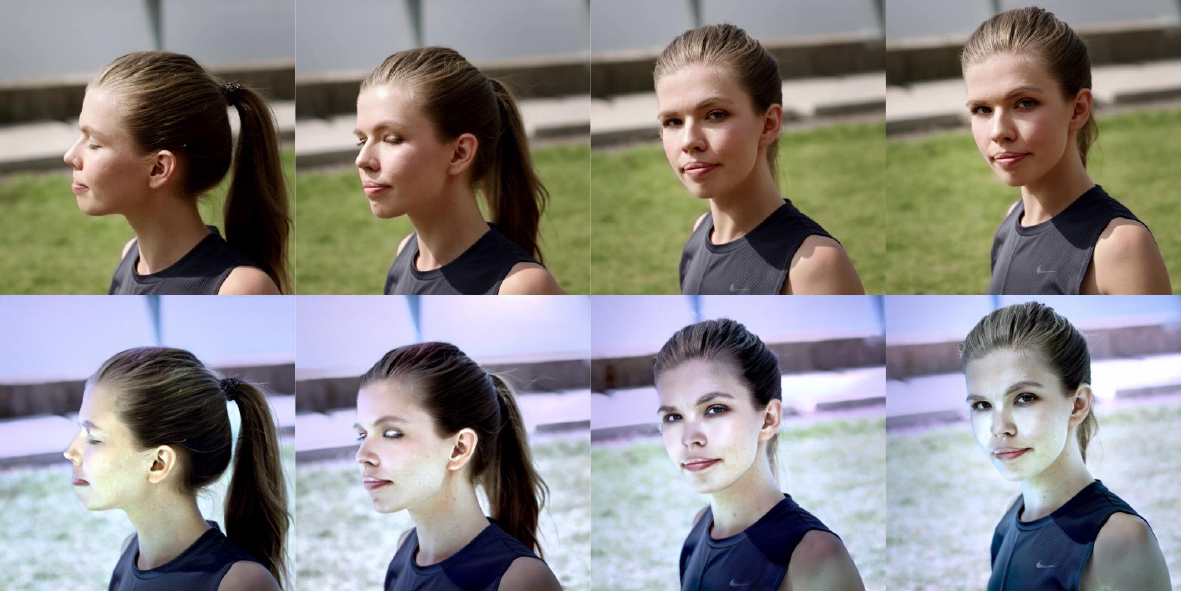}
    \caption{Frame-by-frame results of our method applied to video input. The first row shows the input frames, and the second row presents the corresponding relighting results. The relighting process was guided by the text prompt: ``The magical world covered by snow."}
    \label{fig:application_video}
\end{figure}

\noindent\textbf{Video-based Text2Relight.} We apply our method to video inputs, processing each frame individually. The results are shown in Figure~\ref{fig:application_video}.

\section{User study}
\begin{figure}[t]
    \centering
    \includegraphics[width=0.95\linewidth]{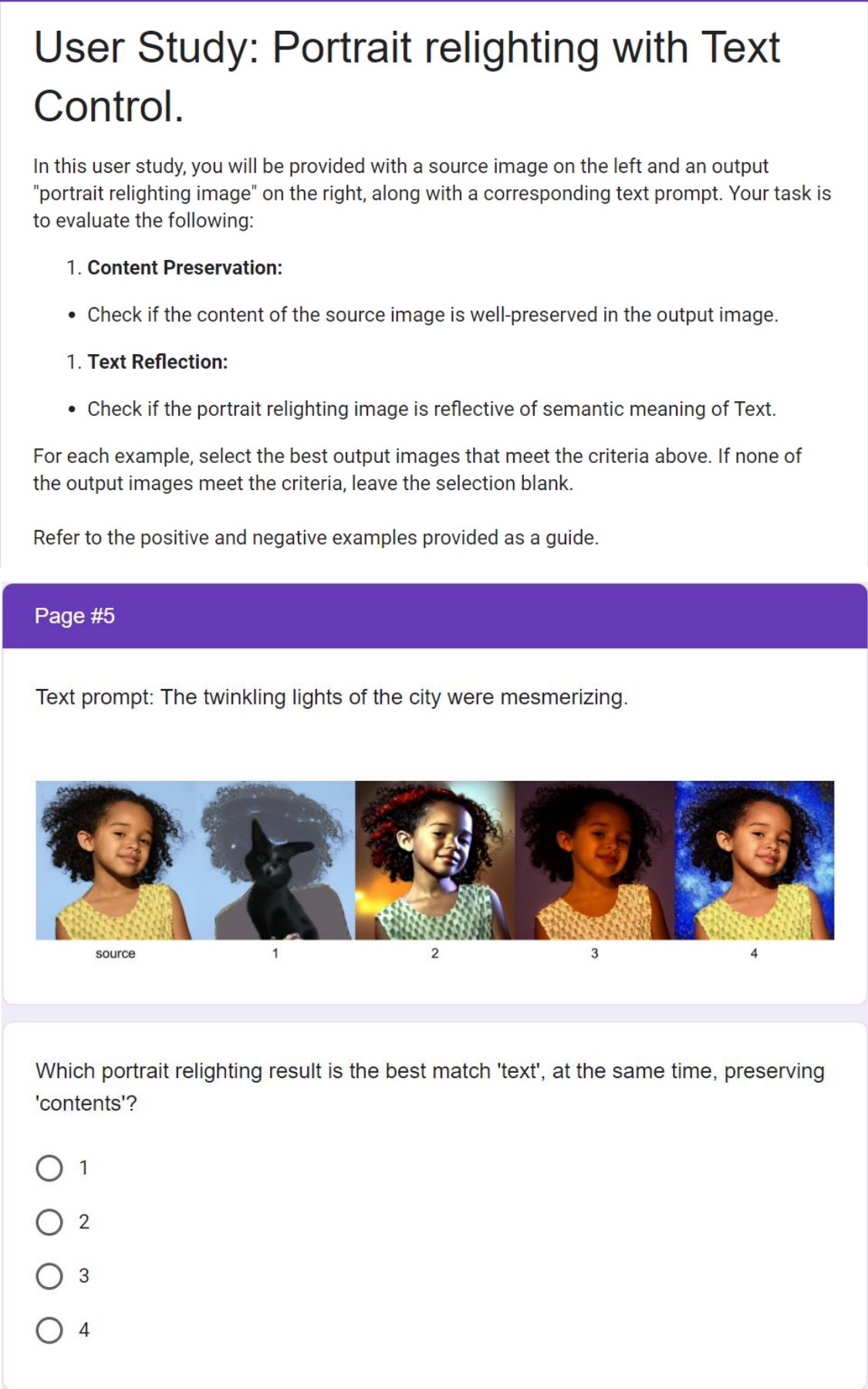}
    \caption{Example of the user study.}
    \label{fig:user_study}
\end{figure}

\begin{figure*}
    \centering
    \includegraphics[width=0.95\linewidth]{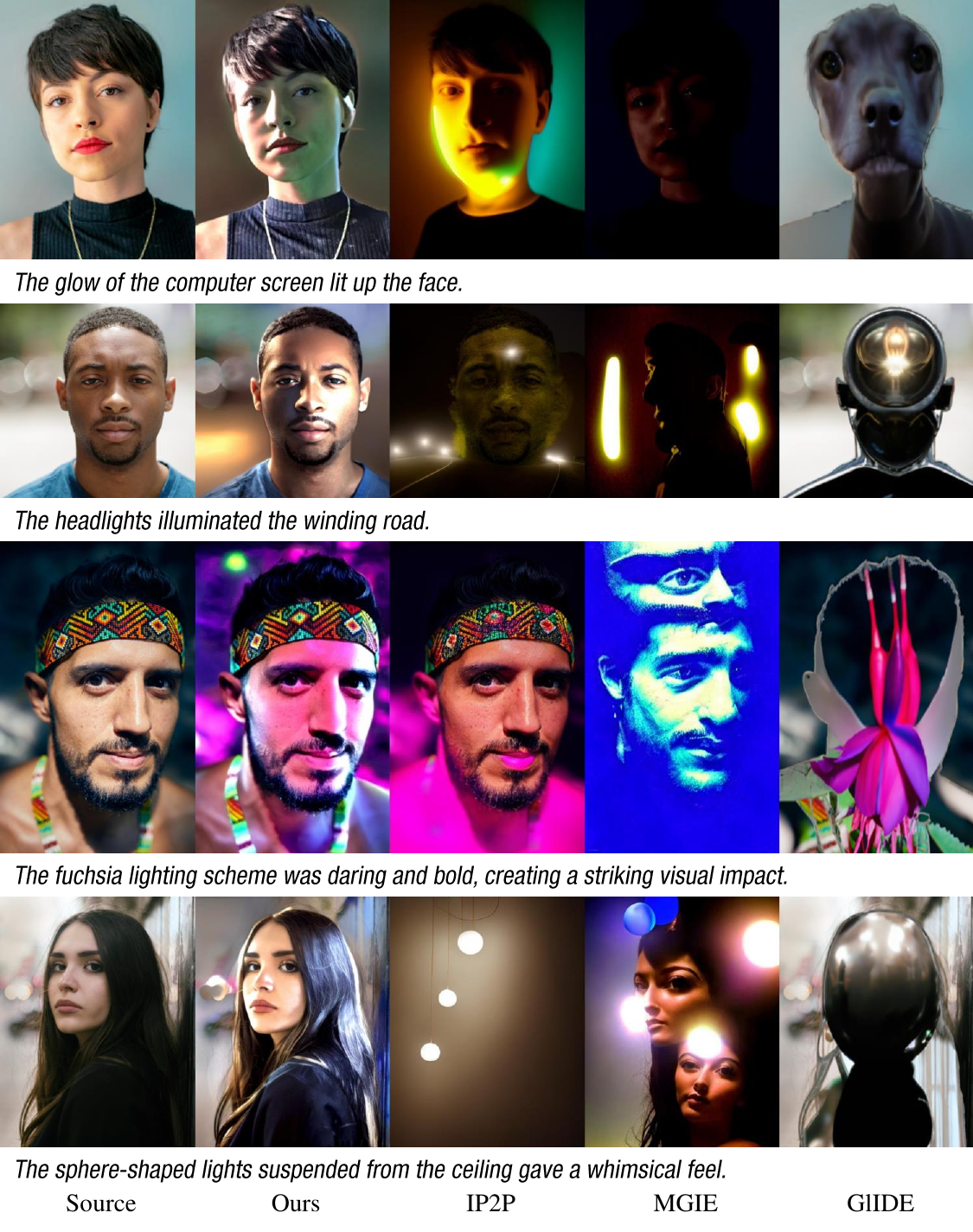}
    \vspace{-5mm}
    \caption{Example images used in the user study.}
    \label{fig:user_study_results1}
\end{figure*}

\begin{figure*}
    \centering
    \includegraphics[width=1\linewidth]{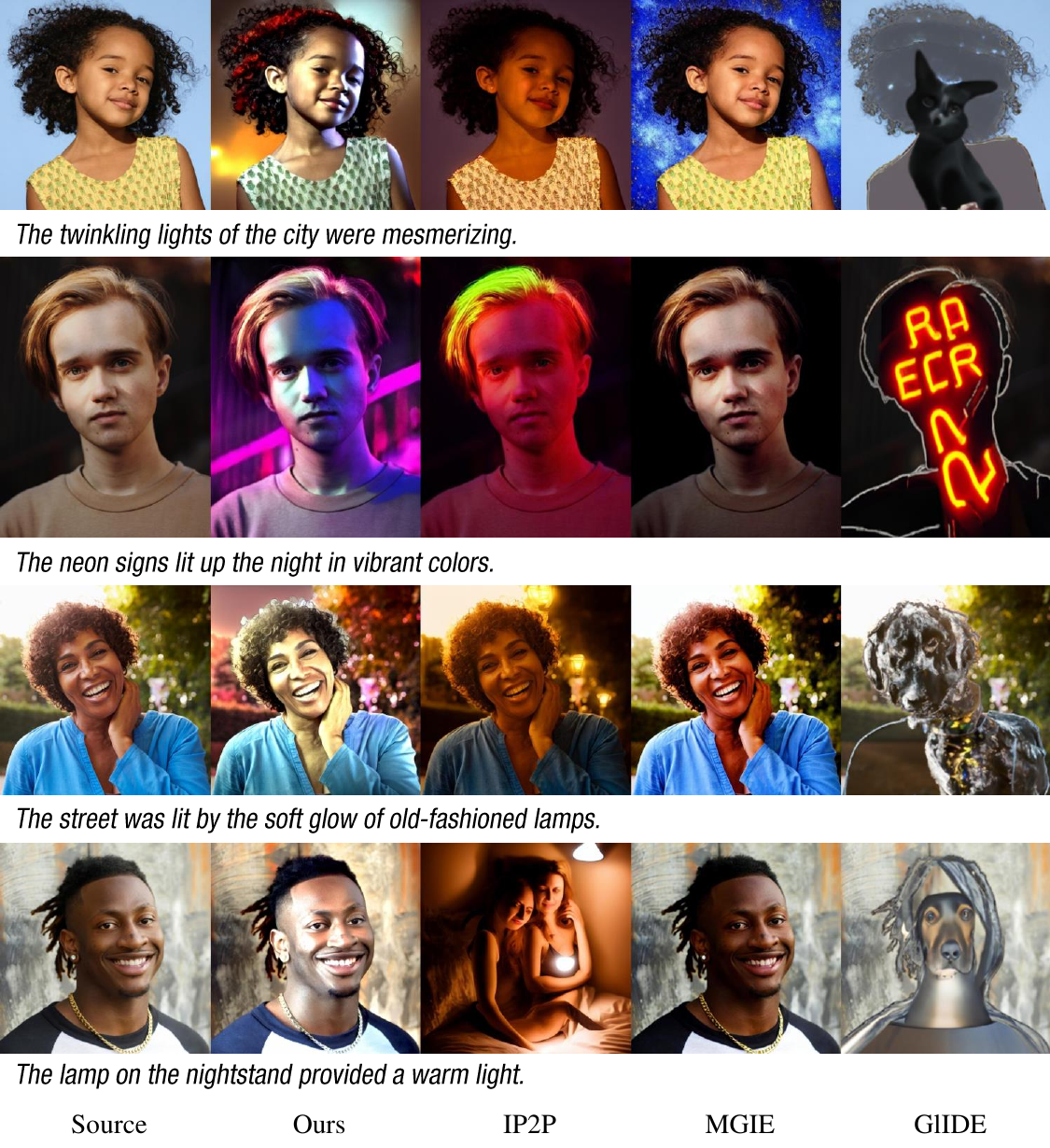}
    \caption{Example images used in the user study.}
    \label{fig:user_study_results2}
\end{figure*}

\begin{figure*}
    \centering
    \includegraphics[width=1\linewidth]{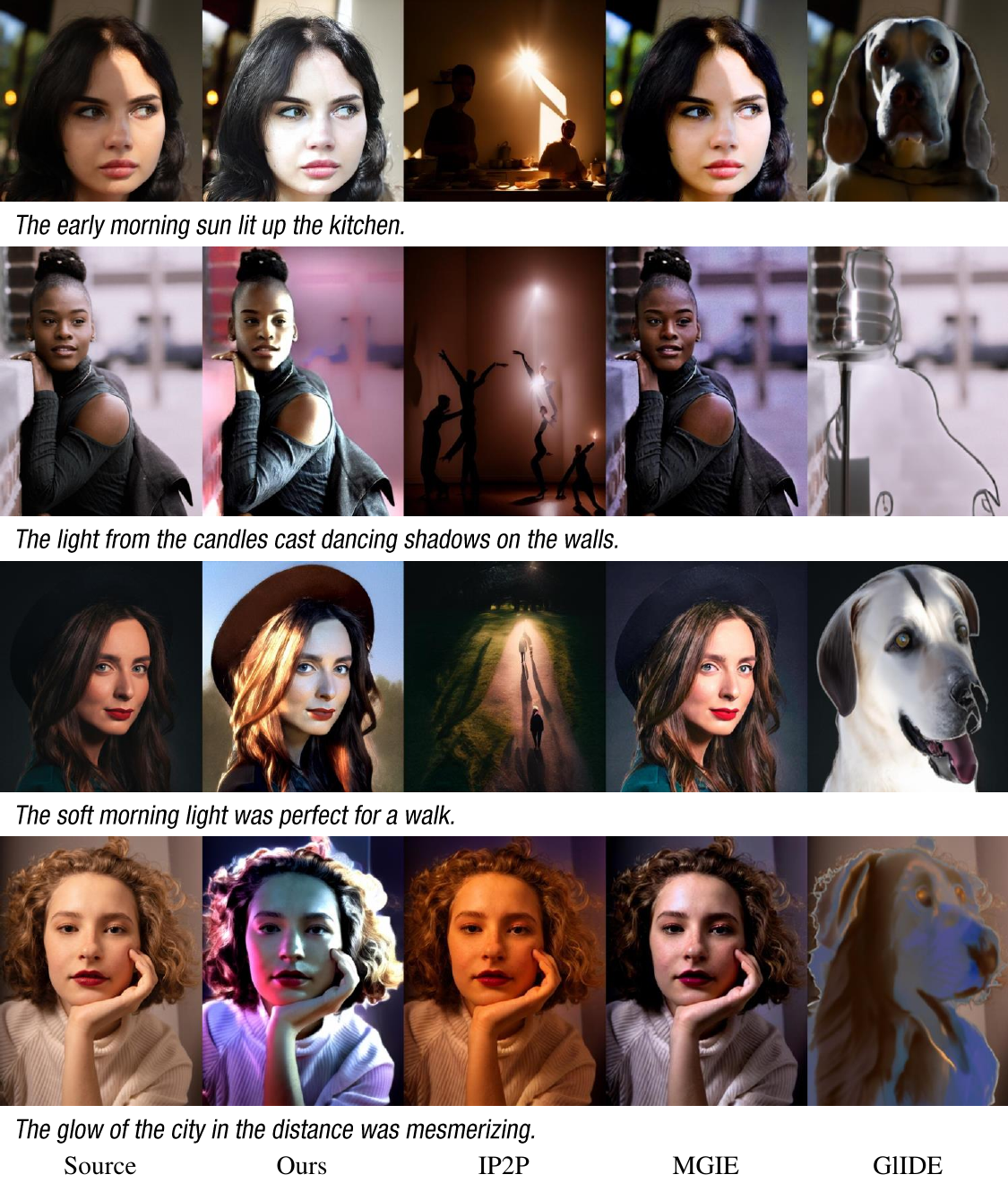}
    \caption{Example images used in the user study.}
    \label{fig:user_study_results3}
\end{figure*}

\begin{figure*}
    \centering
    \includegraphics[width=1\linewidth]{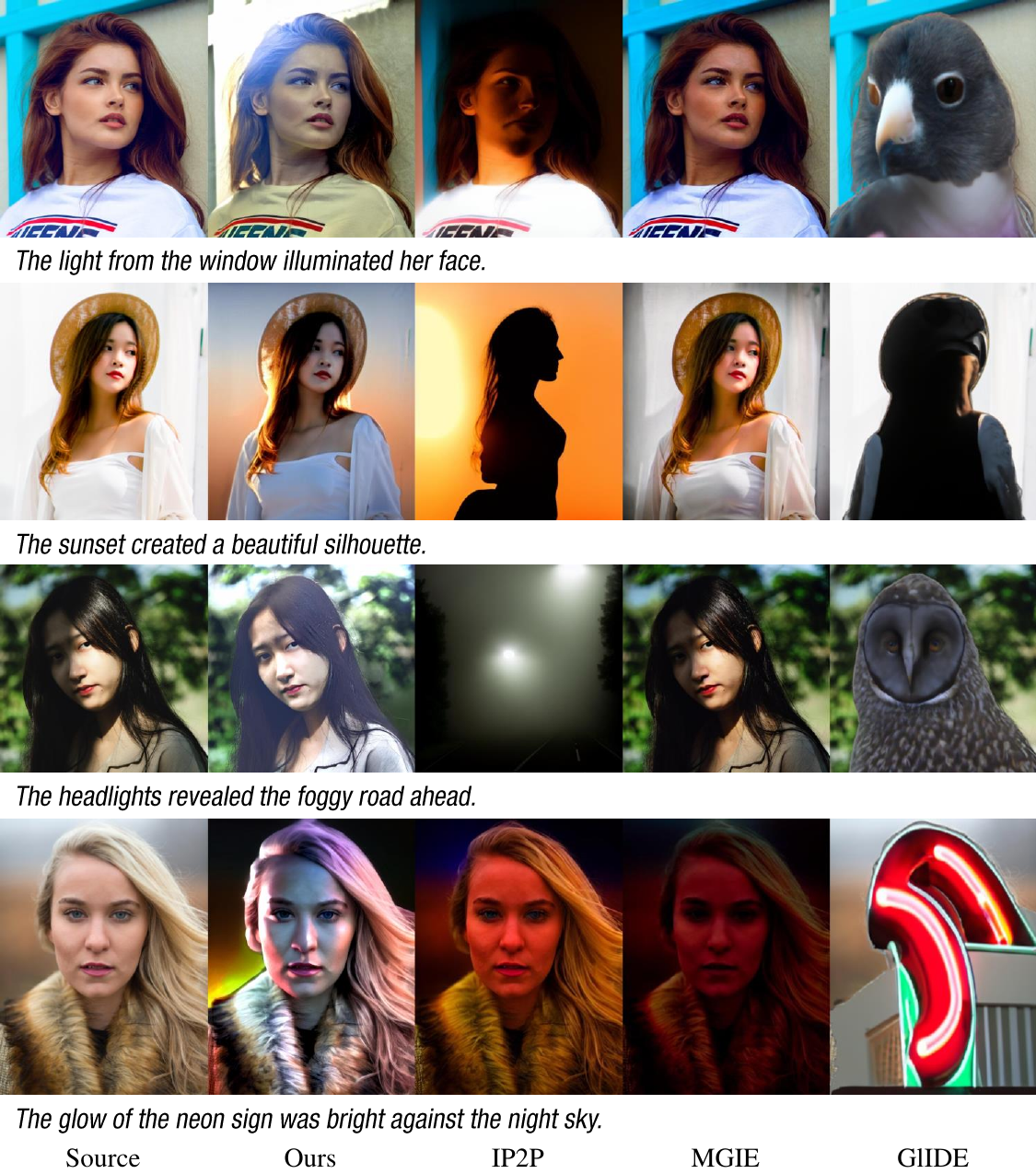}
    \caption{Example images used in the user study.}
    \label{fig:user_study_results4}
\end{figure*}

\begin{figure*}
    \centering
    \includegraphics[width=1\linewidth]{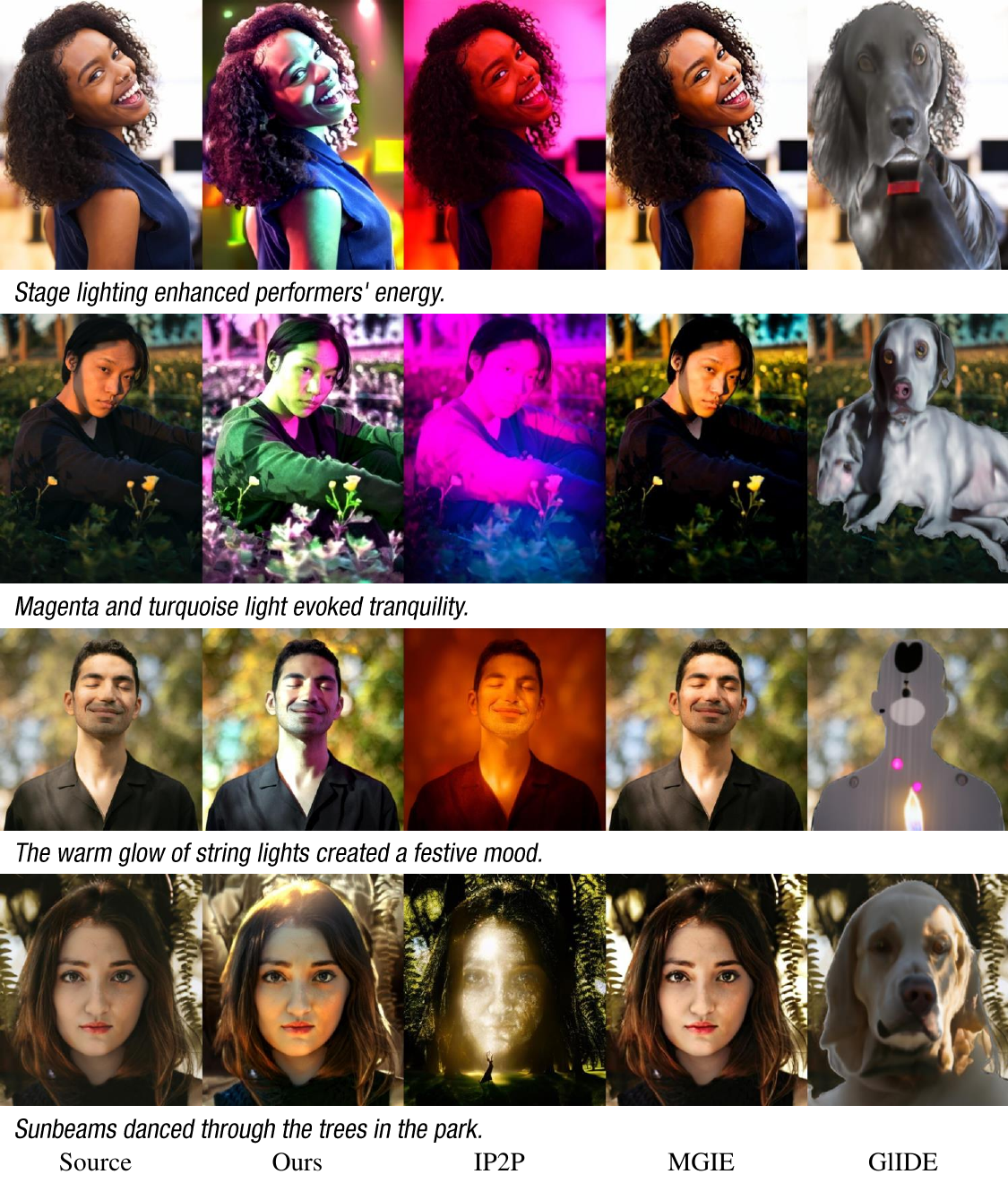}
    \caption{Example images used in the user study.}
    \label{fig:user_study_results5}
\end{figure*}

We conduct a user study as shown in Figure~\ref{fig:user_study}. We ask the users to choose the outputs based on specific criteria: content preservation and text reflection. We show 20 samples and the outputs of four models: ours, IP2P~\cite{brooks2023instructpix2pix}, MGIE~\cite{fu2023guiding}, and GLIDE~\cite{nichol2021glide}, in random order. The image used in the user study can be seen in Figures~\ref{fig:user_study_results1}, \ref{fig:user_study_results2}, \ref{fig:user_study_results3}, \ref{fig:user_study_results4}, and \ref{fig:user_study_results5}.

\section{More results}
\begin{figure*}
    \centering
    \includegraphics[width=0.9\linewidth]{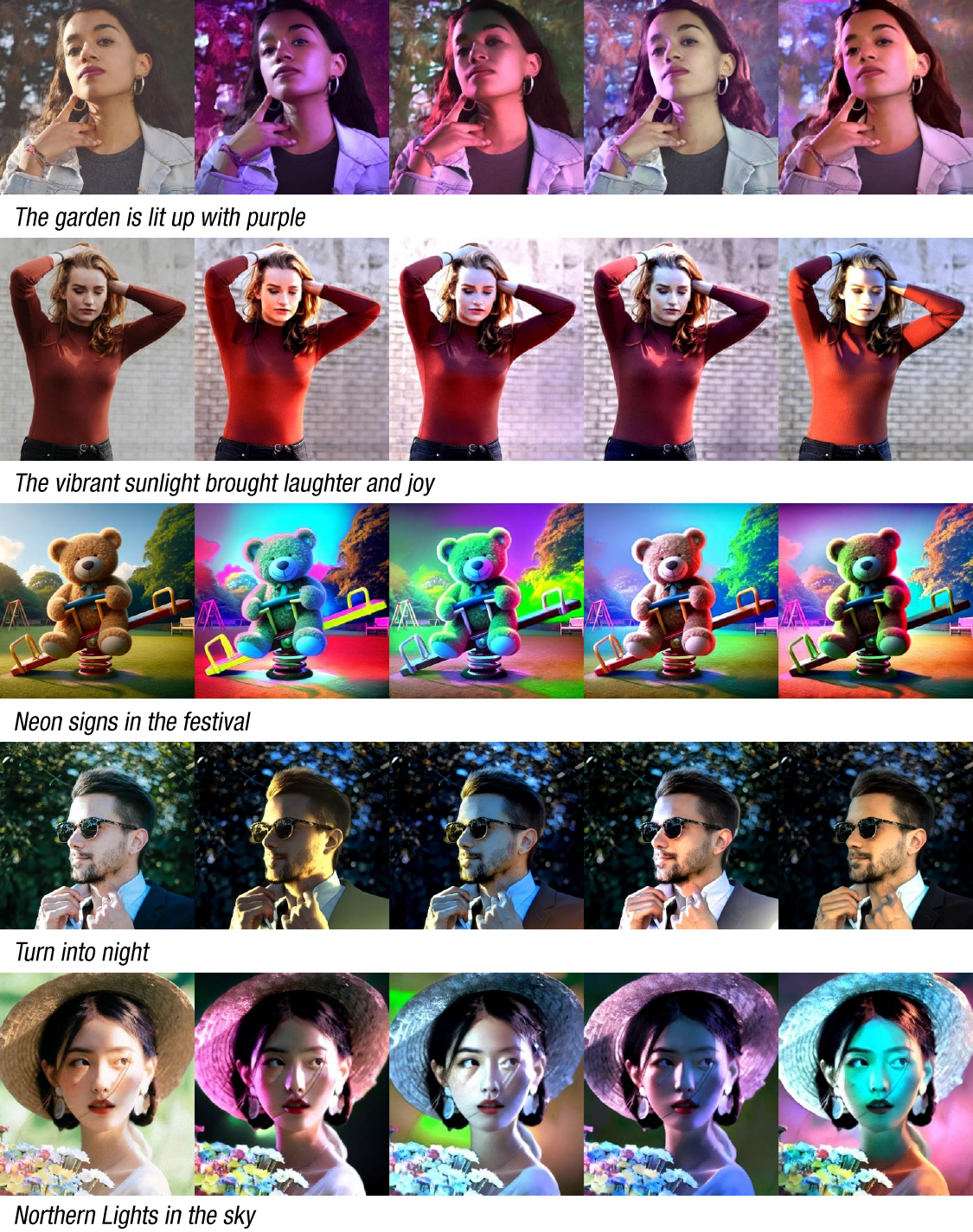}
    \caption{Different results for the same text prompt. The results are obtained by running our model four times. The leftmost image is the source image, and the subsequent four images are the results.}
    \label{fig:multiple_results}
\end{figure*}

\begin{figure*}
    \centering
    \includegraphics[width=1\linewidth]{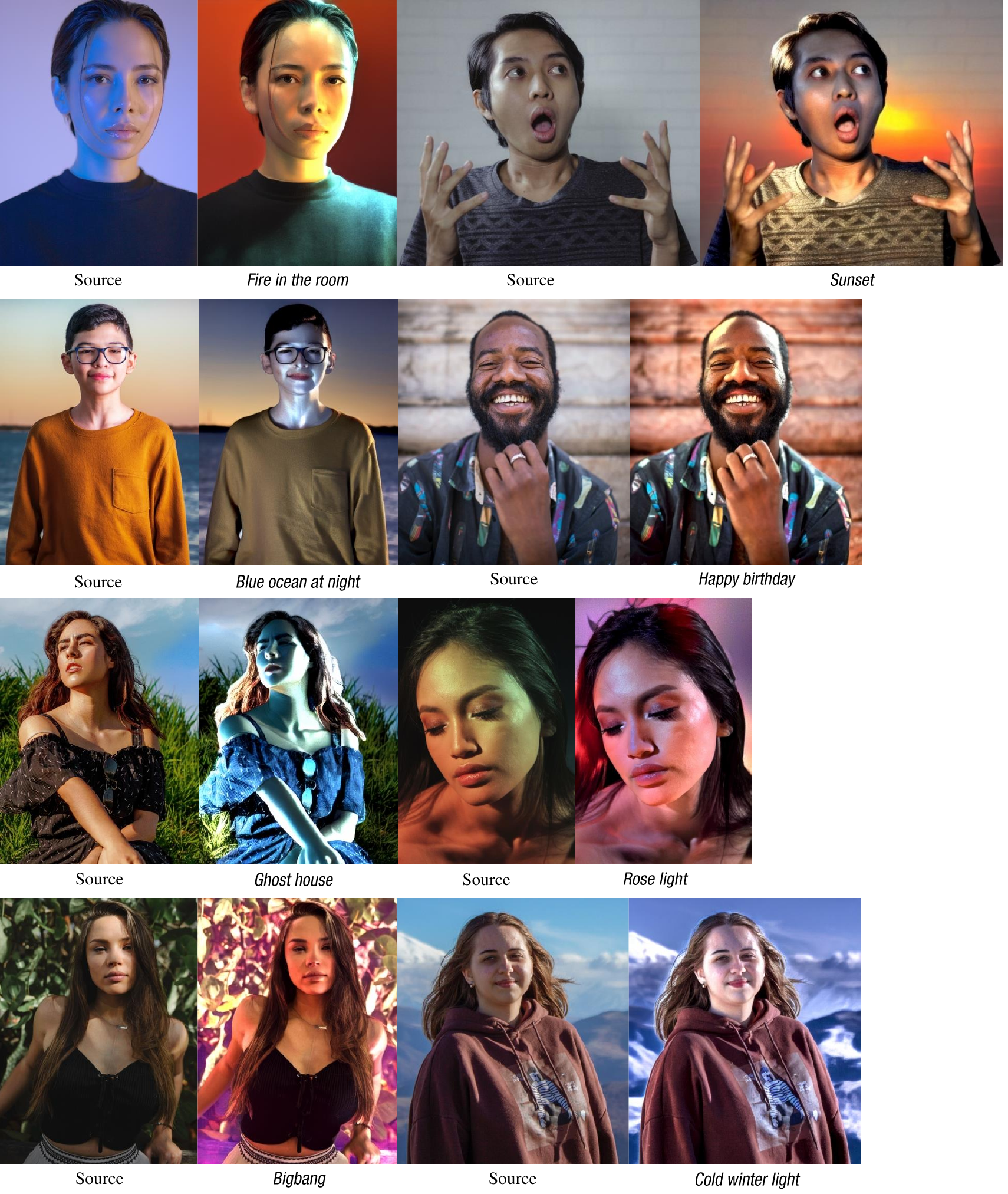}
    \caption{More results.}
    \label{fig:more_results}
\end{figure*}

\begin{figure*}
    \centering
    \includegraphics[width=0.85\linewidth]{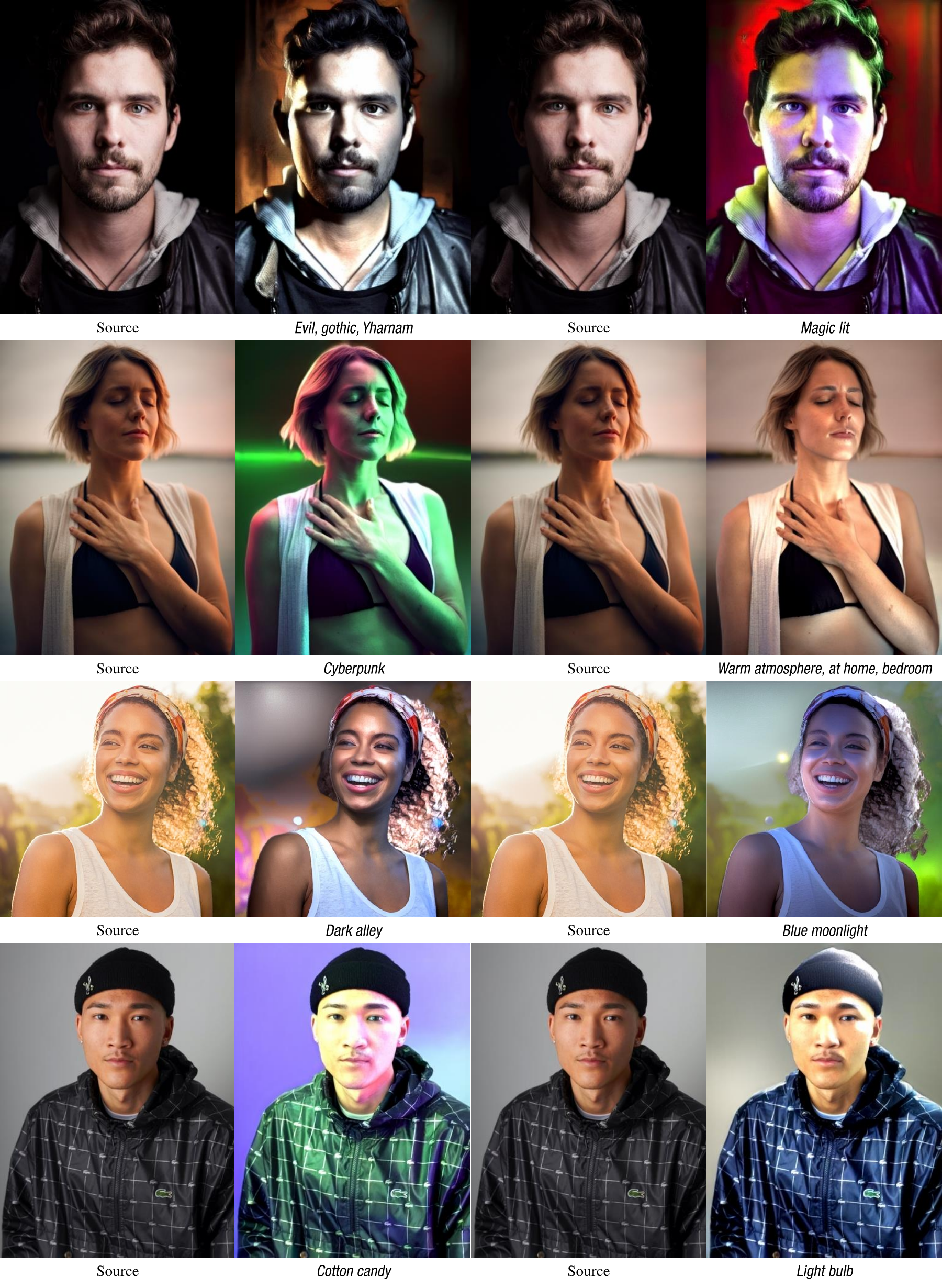}
    \caption{More results.}
    \label{fig:more_results2}
\end{figure*}

\begin{figure*}
    \centering
    \includegraphics[width=1\linewidth]{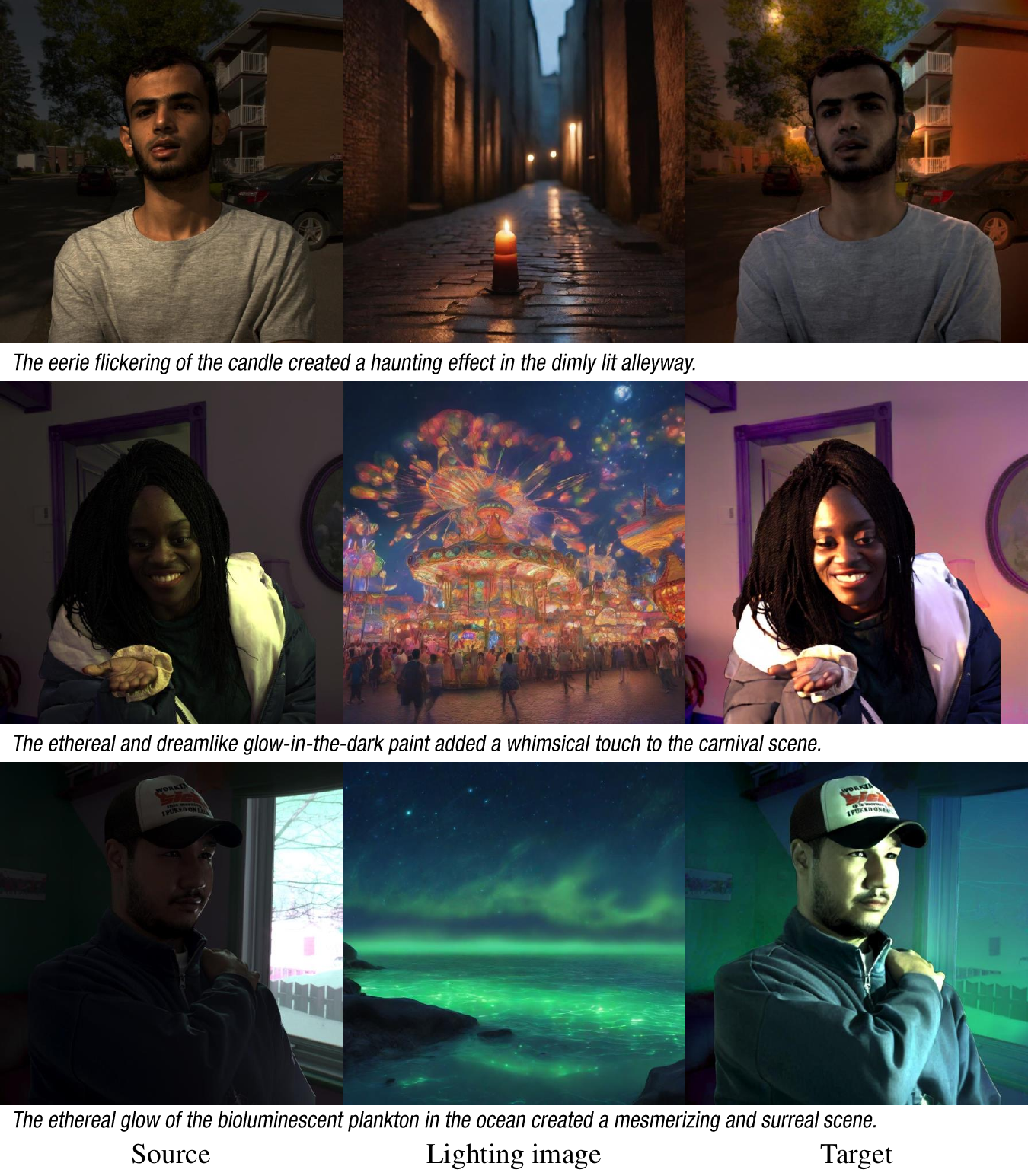}
    \vspace{-5mm}
    \caption{Examples of the generated data.}
    \label{fig:generated_data}
\end{figure*}

\begin{figure*}
    \centering
    \includegraphics[width=1\linewidth]{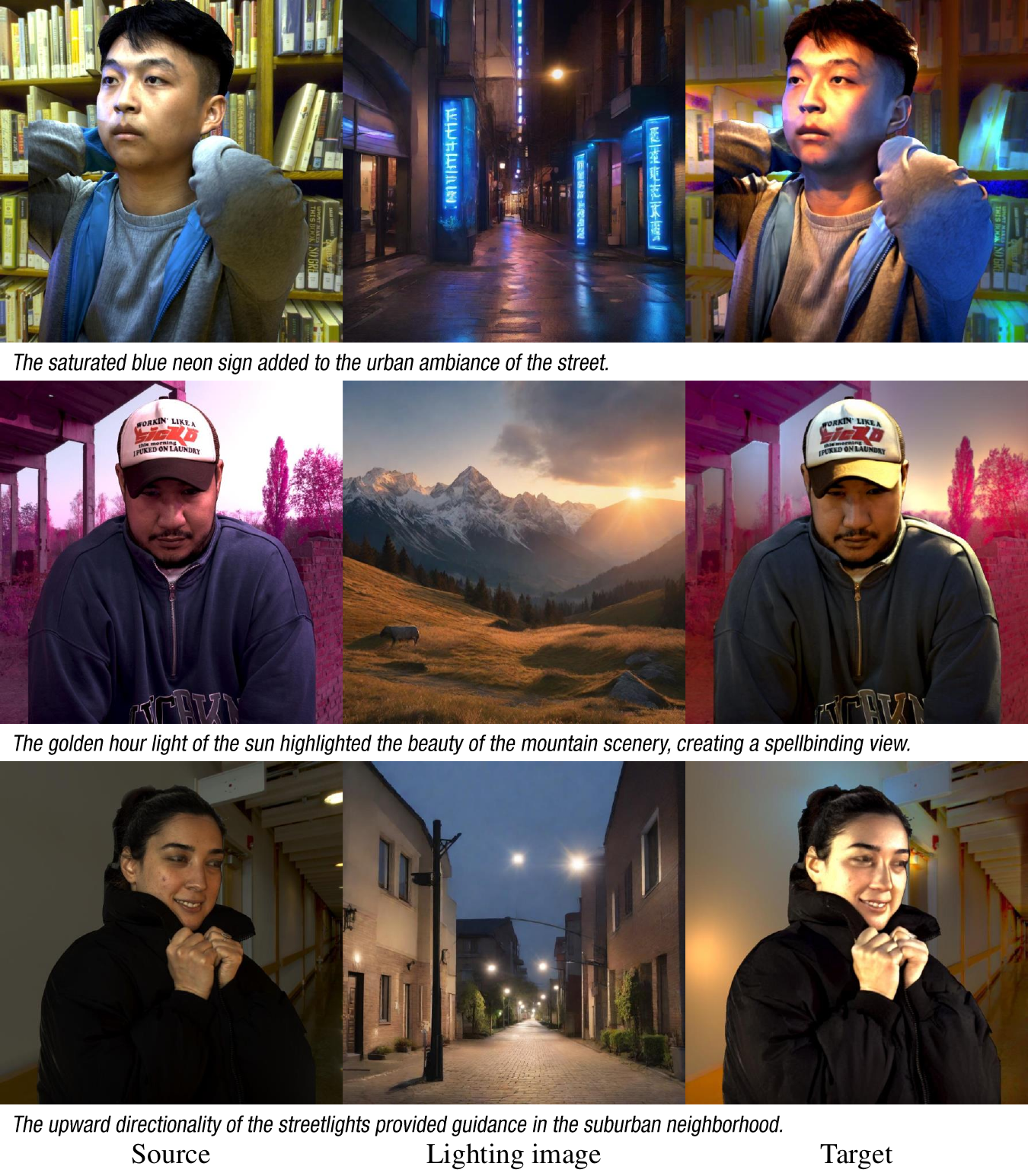}
    \vspace{-5mm}
    \caption{Examples of the generated data.}
    \label{fig:generated_data2}
\end{figure*}

In Figure~\ref{fig:qualitative}, we present the results of applying diverse texts to diverse subjects. In Figure~\ref{fig:multiple_results}, we demonstrate the different results for the same portrait and the same text prompt. We run our model multiple times. In Figure~\ref{fig:more_results} and Figure~\ref{fig:more_results2}, we visualize more results. In Figure~\ref{fig:generated_data} and Figure~\ref{fig:generated_data2}, we visualize our generated data: the source, the lighting image, the target, and the text prompt.

\section{Question Template for LLaVA Score}
Similar to \cite{liu2023gpteval}, our fine-tuned LLAVA~\cite{liu2024visual} requires a specific template to score the image concatenated by a source image and an output image in parallel. We put the following template questions to LLAVA:
\\

\noindent\textbf{Objective:}

Evaluate the effectiveness of the lighting adjustments made to the source image as directed by a text prompt. This task, known as "text2relight," involves modifying the original lighting to align with specified textual instructions.

\noindent\textbf{Evaluation Criteria:}

Lighting Adjustment (1-5) - This criterion assesses how well the lighting changes enhance or alter the atmosphere of the image according to the given directions. The lighting should modify the visual mood without distorting the factual content or the structure of the original image. Points should be deducted for inappropriate lighting changes that obscure details or significantly alter the scene's original elements.

\noindent\textbf{Procedure:}

1. Examine the Adjusted Image: Focus on the lighting changes in the image. Assess how these changes align with the instructions provided in the text prompt, particularly in terms of enhancing or altering the atmosphere.

2. Check for Enhancement: Ensure that the lighting adjustment enhances the overall appearance of the image without distorting the original content and composition.

\noindent\textbf{Text Prompt:}

\{\{Text Prompt\}\}

\noindent\textbf{Evaluation Form (scores ONLY):}

- Lighting Adjustment:
\\

When the text prompt, which is used to generate the output image, is input into \{\{Text Prompt\}\}, LLaVA assigns a score from 1 to 5.

\section{Text Examples}
\subsection{All Categories and Sub-category of Texts}

\noindent\textbf{Time.} Dawn, Sunrise, Evening, Mid-morning, Afternoon, Twilight, Nightfall, Midnight, Nocturnal, Noonday, Morning, Midday, Dusk, Nighttime, Crepuscular, Break of day, Witching hour, Daybreak, Sundown, Gloaming, First light, Last light, Sunset, Day's end, Daytime, Early morning, Late afternoon, Pre-dawn, High noon, Early evening, Late night, Pre-midnight, Post-dawn, Mid-afternoon, Early afternoon, Late morning, Pre-evening, Post-midnight, Early dusk, Late dusk, Pre-twilight

\noindent\textbf{Atmosphere.} Romantic, Mysterious, Dark, Cozy, Enchanted, Gloomy, Ethereal, Whimsical, Surreal, Melancholic, Eerie, Magical, Spellbound, Alluring, Haunting, Dreamlike, Spellbinding, Charming, Moody, Mystical, Enigmatic, Serene, Enchanting, Bewitching, Mystifying, Tranquil, Captivating, Mesmerizing, Shadowy, Luminous

\noindent\textbf{Location.} Indoor, Outdoor, Street, Mansion, Alleyway, Waterfront, Rooftop, Bridge, Plaza, Desert, Cave, Carnival, Market, Courtyard, Balcony, Backyard, Garden, Park, Forest, Lakeside, Mountain, Urban, Rural, Cityscape, Suburban, Wilderness, Seaside, Countryside, Woodland, Beach

\noindent\textbf{Source type.} Sun, LED, Candle, Torch, Fluorescent tube, Neon sign, Firefly, Glow stick, Lightning, Firework, Sparkler, Glow-in-the-dark paint, Lava, Solar, Fire, Oil lamp, Incandescent bulb, Neon, Halogen, Bioluminescent, Gas lamp, Fluorescent, Tungsten, Spotlight, Streetlight, Moonlight, Starlight, Skylight, Aurora, Flashlight

\noindent\textbf{Intensity.} Very bright, Moderate, Faint, Harsh, Brilliant, Gleaming, Dazzling, Flickering, Radiant, Saturated, Softened, Muted, Subtle, Blinding, Soft, Dim, Subdued, Glowing, Luminous, Illuminated, Shimmering, Sparkling, Reflective, Piercing, Intense, Vibrant, Shadowed, Lighted, Fluorescent, Shaded

\noindent\textbf{Temperature.} Warm, Cool, Neutral, Sunlight, Candlelight, Moonlight, Starlight, Twilight, Golden hour, Ice-white, Firelight, Daybreak, Evening glow, Daylight, Ice-blue, Golden, Ivory, Amber, Peach, Sapphire, Violet, Rose, Emerald, Turquoise, Silver, Ruby, Opal, Charcoal, Magenta, Cyan

\noindent\textbf{Directionality.} Direct, Focused, Upward, Scenic, Angular, Off-axis, Radial, Converging, Diverging, Conical, Slanted, Diagonal, Horizontal, Scattered, Indirect, Omni-directional, Diffused, Concentrated, Spread, Multi-directional, Oblique, Refracted, Linear, Vertical, Peripheral, Centralized, Symmetric, Asymmetric, Unidirectional, Cross-directional

\noindent\textbf{Lighting effect.} Shadow creation, Highlight, Blur effect, Halo, Dimming, Accentuating, Radiance, Luminosity, Shimmer, Glint, Gleam, Twinkle, Silhouette, Reflection, Refraction, Lens flare, Sparkle, Glare, Gobo, Lens blur, Light leak, Soft focus, Bokeh, Shadow play, Illumination, Backlighting, Diffusion, Glow, Strobe, Ambient light

\noindent\textbf{Purpose of lighting.} Decorative, Safety, Signaling, Security, Task-specific, Utility, Guidance, Architectural enhancement, Practicality, Illumination, Feature highlighting, Aesthetic enhancement, Festivity, Functional, Mood-setting, Ambient, Accent, Theatrical, Navigation, Festive, Cinematic, Emergency, Energy-saving, Intimate, Directional, Informative, Inviting, Spotlighting, Color-enhancing, Pathway illumination

\noindent\textbf{Emotions.} Happiness, Sadness, Anger, Excitement, Fear, Love, Contentment, Frustration, Anxiety, Peace, Gratitude, Envy, Hope, Jealousy, Disappointment, Surprise, Guilt, Loneliness, Euphoria, Sympathy, Empathy, Pride, Regret, Anticipation, Curiosity, Boredom, Shame, Confidence, Disgust, Awe

\noindent\textbf{Taste.} Flavor, Sweetness, Sourness, Bitterness, Umami, Salty, Savory, Aromatic, Spicy, Tangy, Palate, Preference, Delicacy, Gourmet, Culinary, Gastronomy, Savor, Delectable, Acquired taste, Pungent, Zesty, Flavorful, Bland, Rich, Decadent, Exquisite, Gustatory, Palatable, Crisp, Succulent

\noindent\textbf{Smell.} Fragrance, Aroma, Scent, Odor, Perfume, Stench, Bouquet, Musk, Pungent, Musty, Fresh, Floral, Citrus, Earthy, Woody, Spicy, Sweet, Sour, Bitter, Acrid, Smoky, Minty, Herbal, Fruity, Malodorous, Dank, Lavender, Vanilla, Rosemary, Cinnamon

\noindent\textbf{Sound.} Explosion, Roar, Rumble, Whistle, Screech, Blast, Boom, Chug, Hoot, Howl, Bang, Crash, Shriek, Thunder, Rattle, Squeal, Honk, Clank, Ding, Toot, Clack, Grind, Growl, Buzz, Crack, Pop, Sizzle, Hum, Whir, Click

\noindent\textbf{Touch.} Touch, Texture, Soft, Hard, Smooth, Rough, Sticky, Slippery, Warm, Cold, Moist, Dry, Bumpy, Fuzzy, Prickly, Sharp, Dull, Silky, Velvety, Spongy, Firm, Flexible, Brittle, Squishy, Crisp, Elastic, Fluffy, Grainy, Gritty, Padded

\noindent\textbf{Color.} Black, White, Red, Lime, Blue, Yellow, Cyan, Aqua, Magenta, Fuchsia, Silver, Gray, Maroon, Olive, Green, Purple, Teal, Navy, Dark Red, Brown, Firebrick, Crimson, Tomato, Coral, Indian Red, Light Coral, Dark Salmon, Salmon, Light Salmon, Orange Red, Dark Orange, Orange, Gold, Dark Golden Rod, Golden Rod, Pale Golden Rod, Dark Khaki, Khaki, Yellow Green, Dark Olive Green, Olive Drab, Lawn Green, Chartreuse, Green Yellow, Dark Green, Forest Green, Lime Green, Light Green, Pale Green, Dark Sea Green, Medium Spring Green, Spring Green, Sea Green, Medium Aqua Marine, Medium Sea Green, Light Sea Green, Dark Slate Gray, Dark Cyan, Light Cyan, Dark Turquoise, Turquoise, Medium Turquoise, Pale Turquoise, Aqua Marine, Powder Blue, Cadet Blue, Steel Blue, Corn Flower Blue, Deep Sky Blue, Dodger Blue, Light Blue, Sky Blue, Light Sky Blue, Midnight Blue, Dark Blue, Medium Blue, Royal Blue, Blue Violet, Indigo, Dark Slate Blue, Slate Blue, Medium Slate Blue, Medium Purple, Dark Magenta, Dark Violet, Dark Orchid, Medium Orchid, Thistle, Plum, Violet, Orchid, Medium Violet Red, Pale Violet Red, Deep Pink, Hot Pink, Light Pink, Pink, Antique White, Beige, Bisque, Blanched Almond, Wheat, Corn Silk, Lemon Chiffon, Light Golden Rod Yellow, Light Yellow, Saddle Brown, Sienna, Chocolate, Peru, Sandy Brown, Burly Wood, Tan, Rosy Brown, Moccasin, Navajo White, Peach Puff, Misty Rose, Lavender Blush, Linen, Old Lace, Papaya Whip, Sea Shell, Mint Cream, Slate Gray, Light Slate Gray, Light Steel Blue, Lavender, Floral White, Alice Blue, Ghost White, Honeydew, Ivory, Azure, Snow, Dim Gray, Dark Gray, Silver, Light Gray, Gainsboro, White Smoke.

\noindent\textbf{Shape.} Circle, Square, Triangle, Rectangle, Polygon, Oval, Ellipse, Sphere, Cube, Cylinder, Pyramid, Cone, Prism, Hexagon, Pentagon, Octagon, Diamond, Star, Heart, Crescent, Spiral, Torus, Parallelogram, Rhombus, Trapezoid, Spheroid, Ellipsoid, Lobed, Amorphous, Geometric

\noindent\textbf{Weather.} Rain, Sunny, Cloudy, Windy, Stormy, Snowy, Foggy, Overcast, Drizzly, Freezing, Hot, Cold, Chilly, Warm, Humid, Dry, Muggy, Breezy, Hazy, Icy, Thunderstorms, Light showers, Heavy rain, Scorching, Cool, Frosty, Blustery, Sleet, Mist, Clear night

\noindent\textbf{Universe.} Galaxy, Star, Nebula, Black Hole, Supernova, Comet, Asteroid, Meteor, Satellite, Space, Universe, Cosmology, Quasar, Dark Matter, Dark Energy, Big Bang, Wormhole, Constellation, Orbit, Solar System, Light Year, Red Dwarf, White Dwarf, Pulsar, Event Horizon, Singularity, Cosmic Ray, Aurora, Mercury, Venus, Earth, Mars, Jupiter, Saturn, Uranus, Neptune, Pluto

\noindent\textbf{Light location.} Right, Left, Up, Down, Right bottom, Left bottom, Right up, Left up, Center, Middle, Behind, Front, Corner

\subsubsection{All Action and Position Categories of Texts for Lighting Positioning}

add, incorporate, include, insert, append, remove, delete, eliminate, discard, turn off, shut off, switch off, power down, deactivate, move, relocate, shift, transfer, reposition, turn, alter, transform, modify, convert, farther, more distant, further away, at a greater distance, more remote, closer, nearer, more proximate, at a shorter distance, less distant, brighter, more luminous, more radiant, more brilliant, more illuminated, darker, less illuminated, more shadowy, dimmer, gloomier, more focused, sharper, more concentrated, more pinpointed, more defined, more diffused, more spread out, more dispersed, broader, less concentrated, to the left direction, leftward, to the left side, leftwards, towards the left, to the right direction, rightward, to the right side, rightwards, towards the right, to the down direction, downward, downwards, to the lower side, towards the bottom, to the up direction, upward, upwards, to the upper side, towards the top

\subsubsection{All Color Text Definition and Associated Value for Lighting Positioning}

Black (0, 0, 0), White (255, 255, 255), Red (255, 0, 0), Green (0, 255, 0), Blue (0, 0, 255), Yellow (255, 255, 0), Cyan (0, 255, 255), Magenta (255, 0, 255), Gray (128, 128, 128), Dark Gray (64, 64, 64), Light Gray (192, 192, 192), Maroon (128, 0, 0), Dark Green (0, 128, 0), Navy (0, 0, 128), Olive (128, 128, 0), Teal (0, 128, 128), Purple (128, 0, 128), Brown (165, 42, 42), Pink (255, 192, 203), Orange (255, 165, 0), Gold (255, 215, 0), Silver (192, 192, 192), Sky Blue (135, 206, 235), Forest Green (34, 139, 34), Royal Blue (65, 105, 225), Lime (0, 255, 0), Turquoise (64, 224, 208), Violet (238, 130, 238), Indigo (75, 0, 130), Beige (245, 245, 220), Salmon (250, 128, 114), Khaki (240, 230, 140), Chocolate (210, 105, 30), Aqua (0, 255, 255), Hot Pink (255, 105, 180), Tomato (255, 99, 71), Cornflower Blue (100, 149, 237), Slate Gray (112, 128, 144), Dark Slate Gray (47, 79, 79), Firebrick (178, 34, 34), Dark Orchid (153, 50, 204), Peru (205, 133, 63), Saddle Brown (139, 69, 19), Sea Green (46, 139, 87), Dark Olive Green (85, 107, 47), Rosy Brown (188, 143, 143), Lavender (230, 230, 250), Dark Salmon (233, 150, 122), Medium Aquamarine (102, 205, 170), Medium Purple (147, 112, 219), Pale Green (152, 251, 152), Steel Blue (70, 130, 180), Dark Khaki (189, 183, 107), Light Coral (240, 128, 128), Dark Cyan (0, 139, 139), Medium Violet Red (199, 21, 133), Slate Blue (106, 90, 205), Medium Sea Green (60, 179, 113), Light Sky Blue (135, 206, 250), Lemon Chiffon (255, 250, 205), Medium Slate Blue (123, 104, 238), Pale Turquoise (175, 238, 238), Light Salmon (255, 160, 122), Light Steel Blue (176, 196, 222), Medium Orchid (186, 85, 211), Light Green (144, 238, 144), Burlywood (222, 184, 135), Dark Turquoise (0, 206, 209), Cadet Blue (95, 158, 160), Light Pink (255, 182, 193), Medium Spring Green (0, 250, 154), Dark Sea Green (143, 188, 143), Light Slate Gray (119, 136, 153), Deep Pink (255, 20, 147), Pale Violet Red (219, 112, 147), Light Blue (173, 216, 230), Powder Blue (176, 224, 230), Orchid (218, 112, 214), Light Yellow (255, 255, 224), Medium Blue (0, 0, 205), Corn Silk (255, 248, 220), Dark Orange (255, 140, 0), Light Goldenrod Yellow (250, 250, 210), Medium Turquoise (72, 209, 204), Dark Goldenrod (184, 134, 11), Peach Puff (255, 218, 185), Lawn Green (124, 252, 0), Light Cyan (224, 255, 255), Dark Red (139, 0, 0), Olive Drab (107, 142, 35), Dark Violet (148, 0, 211), Pale Goldenrod (238, 232, 170), Thistle (216, 191, 216), Indian Red (205, 92, 92), Dark Slate Blue (72, 61, 139), Dark Gray (169, 169, 169), Medium Slate Blue (123, 104, 238), Light Grey (211, 211, 211)


\begin{figure*}
    \centering
    \includegraphics[width=0.93\linewidth]{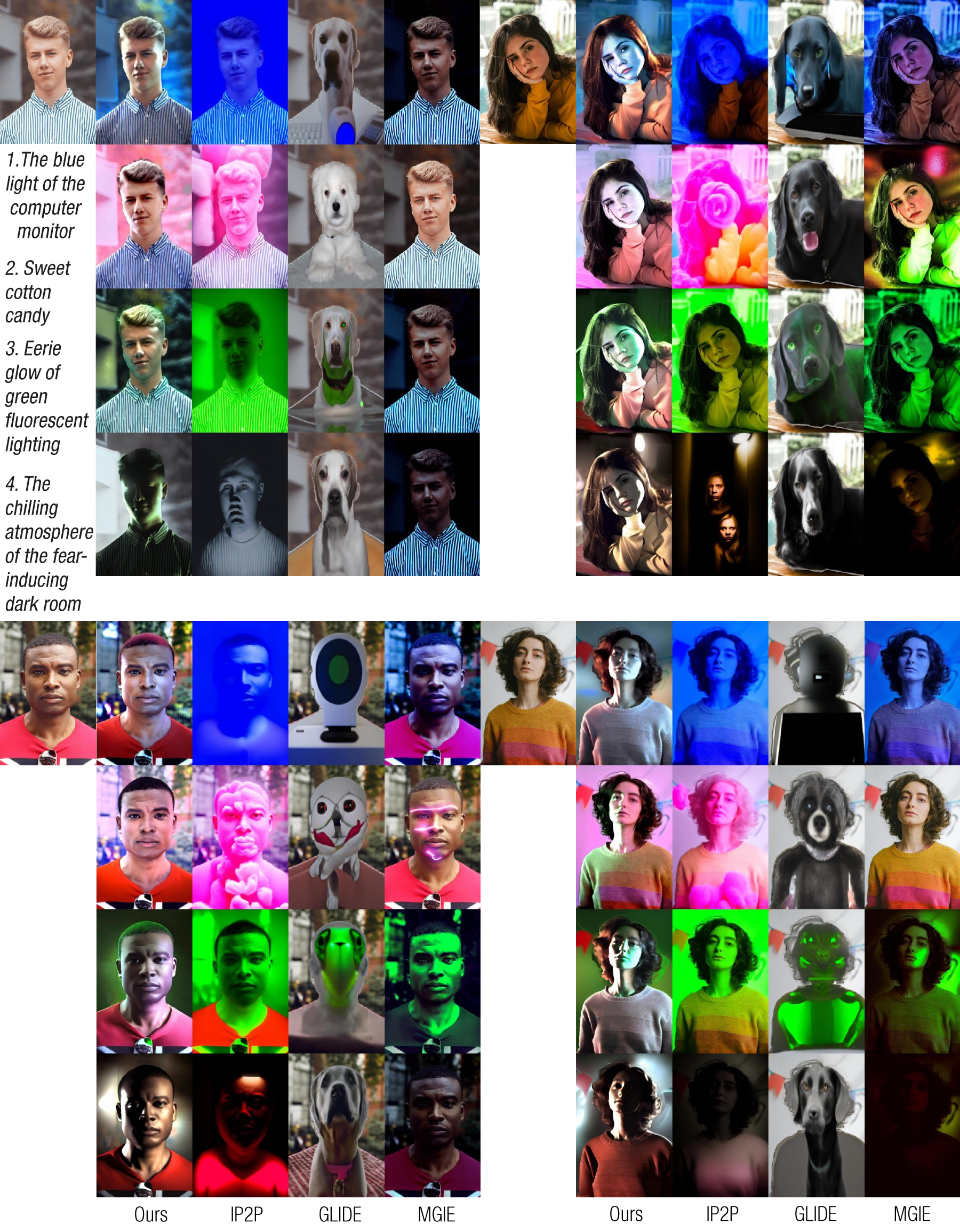}
    \caption{Qualitative comparison. We compare ours with IP2P~\cite{brooks2023instructpix2pix}, GLIDE~\cite{nichol2021glide}, and MGIE~\cite{fu2023guiding}. In the 2x2 grid, the top-left image in each grid is the source image, and each row shows the results obtained using text prompts 1 through 4.}
    \label{fig:qualitative}
\end{figure*}

\end{document}